\pdfoutput=1

\documentclass[11pt]{article}

\usepackage{EMNLP2022}

\usepackage{times}
\usepackage{latexsym}
\usepackage{amssymb}

\usepackage[T1]{fontenc}
\usepackage{multirow}
\usepackage{graphicx}
\usepackage{CJKutf8}
\usepackage[utf8]{inputenc}
\usepackage{amsmath}

\usepackage{microtype}
\usepackage{booktabs}
\usepackage{inconsolata}
\usepackage{lscape}
\usepackage{longtable}
\usepackage{tablefootnote}

\usepackage[russian,main=english]{babel}
\babelprovide[import]{russian}

\usepackage{CJKutf8} 

\newcommand{\name}[0]{\textsc{Par3}}

\newcommand{\red}[1]{\textbf{\textcolor{red}{#1}}}
\newcommand{\violet}[1]{\textbf{\textcolor{violet}{#1}}}
\newcommand{\teal}[1]{\textbf{\textcolor{teal}{#1}}}
\newcommand{\orange}[1]{\textbf{\textcolor{orange}{#1}}}
\newcommand{\blue}[1]{\textbf{\textcolor{blue}{#1}}}
\newcommand\blankfootnote[1]{%
  \let\thefootnote\relax\footnotetext{#1}%
  \let\thefootnote\svthefootnote%
}

\newcommand{\titlestr}{Exploring Document-Level Literary Machine Translation \\ with Parallel Paragraphs from World Literature}

\title{\titlestr}

\author{Katherine Thai$^{\bigstar\diamondsuit}$ \quad Marzena Karpinska$^{\bigstar\diamondsuit}$ \quad Kalpesh Krishna$^\diamondsuit$ \quad William Ray$^\diamondsuit$\\  {\bf Moira Inghilleri}$^\spadesuit$ \quad {\bf John Wieting}$^{\clubsuit}$  \quad {\bf Mohit Iyyer}$^{\diamondsuit}$\\\\
$^\diamondsuit$Manning College of Information and Computer Sciences, UMass Amherst\\ $^\spadesuit$Department of Languages, Literatures, and Cultures; UMass Amherst\\
$^\clubsuit$Google Research\\
\texttt{\{kbthai,mkarpinska,kalpesh,miyyer\}@cs.umass.edu}\\ \texttt{minghilleri@complit.umass.edu}, \texttt{jwieting@google.com}}

\begin{document}
\maketitle
\begin{abstract}
    Literary translation is a culturally significant task, but it is bottlenecked by the small number of qualified literary translators relative to the many untranslated works published around the world. Machine translation (MT) holds potential to complement the work of human translators by improving both training procedures and their overall efficiency. Literary translation is less constrained than more traditional MT settings since translators must balance meaning equivalence, readability, and critical interpretability in the target language. This property, along with the complex discourse-level context present in literary texts, also makes literary MT more challenging to computationally model and evaluate. To explore this task, we collect a dataset (\name) of non-English language novels in the public domain, each aligned at the paragraph level to both human and automatic English translations. Using \name, we discover that expert literary translators prefer reference human translations over machine-translated paragraphs at a rate of 84\%, while state-of-the-art automatic MT metrics do not correlate with those preferences. The experts note that MT outputs contain not only mistranslations, but also discourse-disrupting errors and stylistic inconsistencies. To address these problems, we train a post-editing model whose output is preferred over normal MT output at a rate of 69\% by experts. We publicly release  \name\  to spur future research into literary MT.\footnote{\url{https://github.com/katherinethai/par3/}}
    \blankfootnote{$\bigstar$ Authors contributed equally.}
\end{abstract}
\section{Introduction}
\label{sec:introduction}

While the quality of machine translation (MT) systems has greatly improved with recent advances in modeling and dataset collection, the application of these new technologies to the task of automatically translating \emph{literary} text (e.g., novels, short stories) has remained limited to  small-scale studies~\citep{genzel-etal-2010-poetic,jones-irvine-2013-un,toral2018post}. 
Translating literary works differs from translating standard MT corpora (e.g., news articles or parliamentary proceedings) in several key ways. For one, it is much more difficult to evaluate. 
The techniques\footnote{Many terms have been employed by translation scholars to refer to various operations used by translators \citep{Chesterman2017-strat-papaers}. Here, we employ the term ``techniques'' argued for by \citet*{Molina2004-ms} and recently used in the field of NLP \citep{zhai-etal-2018-construction, zhai-etal-2020-building}.} used by literary translators differ fundamentally from those applied in more standard MT domains (see Table \ref{tab:strategies} in the Appendix). Literary translators have the freedom (or burden) of both semantic and critical interpretation, as they must solve the problem of \textit{equivalence}, often beyond the word level 
\citep{neubert1983discourse,Baker2018-cg,Baker2021-qr-encyclopedia-transstudy}. The task of conveying an author's ideas highlights yet another difference between literary and traditional MT: \emph{document-level} context is especially critical for the literary domain due to the presence of complex discourse structure, rendering the typical sentence-level MT pipeline insufficient for this task~\citep{voigt2012towards,taivalkoski2019free}.

In this work, we seek to understand how both state-of-the-art MT systems and MT evaluation metrics fail in the literary domain, and we also leverage large pretrained language models to improve literary MT. To facilitate our experiments, we introduce \name, a large-scale dataset to study \emph{paragraph-level literary translation} into English. \name\ consists of 121K paragraphs taken from 118 novels originally written in a non-English language, where each paragraph is aligned to multiple human-written English translations of that paragraph as well as a machine-translated paragraph produced by Google Translate (see Table~\ref{tab:par3_example}).

We show that MT evaluation metrics such as \textsc{Bleu} and \textsc{Bleurt} are not effective for literary MT. In fact, we discover that two of our tested metrics (\textsc{Bleu} and the document-level \textsc{BlonDe}) show a preference for Google Translate outputs over reference translations in \name. In reality, MT outputs are much worse than reference translations: our human evaluation reveals that professional translators prefer reference translations at a rate of \textbf{85\%}. 

While the translators in our study identified overly literal translations and discourse-level errors (e.g., coreference, pronoun consistency) as the main faults of modern MT systems, a monolingual human evaluation comparing human reference translations and MT outputs reveals additional hurdles in readability and fluency. To tackle these issues, we fine-tune GPT-3~\citep{brown2020language} on an automatic post-editing task in which the model attempts to transform an MT output into a human reference translation. Human translators prefer the post-edited translations at a rate of 69\% and also observe a lower incidence of the above errors.

Overall, we identify critical roadblocks in evaluation towards meaningful progress in literary MT, and we also show through expert human evaluations that pretrained language models can improve the quality of existing MT systems on this domain. We release \name\ to spur more meaningful future research in literary MT.

\section{The \name\ Dataset: Parallel Paragraph-Level Paraphrases}

\begin{table}[ht]
\renewcommand{\arraystretch}{1.15}
\scriptsize
  \centering
  \begin{tabular}{l|c|c|c}
    \hline
    \textbf{Src lang} & \textbf{\#texts} & \textbf{\#src paras} & \textbf{sents/para}\\
    \hline
    French (\textit{fr}) & 32  & 50,070 & 2.7\\

    Russian (\textit{ru}) & 27  & 36,117 & 3.3\\ 

    German (\textit{de}) & 16  & 9,170  & 4.3\\ 

    Spanish (\textit{es}) & 1  & 3,279 & 2.0\\

    Czech (\textit{cs}) & 4  & 2,930 & 3.0\\
    
    Norwegian (\textit{no}) & 2  & 2,655  & 3.4\\  
    
    Swedish (\textit{sv}) & 3  & 2,620  & 3.2 \\ 
    
    Portuguese (\textit{pt}) & 4  & 2,288 & 3.7\\  
    
    Italian (\textit{it}) & 2  & 1,931 & 2.6 \\ 

    Japanese (\textit{ja}) & 9  & 1,857 & 4.4\\  

    Bengali (\textit{bn}) & 2  & 1,499 & 3.3 \\ 
        
    Tamil (\textit{ta}) & 1  & 1,489 & 3.1\\

    Danish (\textit{da}) & 1 & 1,384 & 3.6 \\ 

    Chinese\tablefootnote{The Chinese texts in \name\ were written in Classical Chinese, an archaic and very different form of the language currently used today.} (zh) & 7  & 1,320  & 8.8\\  

    Dutch (\textit{nl}) & 1  & 963 & 3.4\\

    Hungarian (\textit{hu}) & 1  & 892 & 3.7\\
    
    Polish (\textit{pl}) & 1  & 399 & 3.9\\

    Sesotho (\textit{st}) & 1  & 374 & 4.2\\
    
    Persian (\textit{fa}) & 1  & 148 & 4.2\\
    \hline
    \textbf{All} & 118 & 121,385 & 3.2\\
    \hline
  \end{tabular}
  \caption{
      Corpus statistics for Version 2 of \name\ by each of the 19 source languages. The average number of sentences per paragraph refers to only the English human and Google translations of the source paragraphs. We did not count tokens or sentences for source paragraphs because of the lack of a reliable tokenizer and sentence segmenter for all source languages. 
  }
  \label{tab:basic_stats}
\end{table}











    
    
    
    
    


To study literary MT, we collect a dataset of \textbf{par}allel \textbf{par}agraph-level \textbf{par}aphrases (\name) from public domain non-English-language (\textit{source}) novels with their corresponding English translations generated by both humans and Google Translate. \name\ is a step up in both scale and linguistic diversity compared to prior studies in literary MT, which generally focus on one novel~\citep{toral2018post} or a small set of poems or short stories~\citep{jones-irvine-2013-un}. \name\ contains at least two human translations for every source paragraph (Table~\ref{tab:par3_example}). In Table \ref{tab:basic_stats}, we report corpus statistics by the 19 unique source languages\footnote{Languages in \name\ represent different language families (Romance, Germanic, Slavic, Japonic, Sino-Tibetan, Iranian, Dravidian, Ugric, and Bantu), with different morphological traits (synthetic, fusional, agglutinative), and use different writing systems (Latin alphabet, Cyrillic alphabet, Bengali script, Persian alphabet, Tamil script, Hanzi, and Kanji/Hiragana/Katakana).} represented in \name. \name\ was curated in four stages: selection of source texts, machine translation of source texts, paragraph alignment, and final filtering. This process closely resembles the paraphrase mining methodology described by~\citet{barzilay-mckeown-2001-extracting}; the major distinctions are (1) our collection of literary works that is $\sim20$ times the size of the previous work, (2) our inclusion of the aligned source text to enable translation study, and (3) our alignment at the paragraph, not sentence, level. In this section, we describe the data collection process and disclose choices we made during curation of Version 1 of \name. See Section \ref{sec: versions} in the Appendix for more details on the different versions of \name.  

\begin{table*}[]
    \scriptsize
    \begin{tabular}{p{3.6cm}p{3.6cm}p{3.6cm}p{3.6cm}}
        \multicolumn{4}{p{15cm}}{\texttt{SRC (ru):} \begin{otherlanguage}{russian}— Извините меня: я, увидевши издали, как вы вошли в лавку, решился вас побеспокоить. Если вам будет после свободно и по дороге мимо моего дома, так сделайте милость, зайдите на малость времени. Мне с вами нужно будет переговорить\end{otherlanguage}}

        \vspace{.05cm}
        
        \\
        
        \hline
        
        \vspace{.05cm}
        
        \texttt{GTr}: ``Excuse me; seeing from a distance how you entered the shop, I decided to disturb you. If you will be free after and on the way past my house, so do yourself a favour, stop by for a little time. I will need to speak with you. &
        
        \vspace{.05cm}
        \texttt{HUM1}: ``Pardon me, I saw you from a distance going into the shop and ventured to disturb you. If you will be free in a little while and will be passing by my house, do me the favour to come in for a few minutes. I want to have a talk with you.'' & 
        
        \vspace{.05cm}
        \texttt{HUM2}: ``I saw you enter the shop,'' he said, ``and therefore followed you, for I have something important for your ear. Could you spare me a minute or two?'' & 
        
        \vspace{.05cm}
        \texttt{HUM3}: `Excuse me: I saw you from far off going into the shop, and decided to trouble you. If you’re free afterwards and my house is not out of your way, kindly stop by for a short while. I must have a talk with you.''
        
        \vspace{.3cm}
        \\
   
        \multicolumn{4}{p{15cm}}{\texttt{SRC (st):} Ho bile jwalo ho fela ha Chaka, mora wa Senzangakhona. Mazulu le kajeno a bokajeno ha a hopola kamoo a kileng ya eba batho kateng, mehleng ya Chaka, kamoo ditjhaba di neng di jela kgwebeleng ke ho ba tshoha, leha ba hopola borena ba bona bo weleng, eba ba sekisa mahlong, ba re: "Di a bela, di a hlweba! Madiba ho pjha a maholo!" }

        \vspace{.1cm}
        
        \\
        \hline
        \multicolumn{4}{c}{\begin{tabular}{p{4.8cm}p{4.8cm}p{4.8cm}}
         \vspace{.05cm}
         \texttt{GTr}:  Such was the end of Chaka, son of Senzangakhona. The Zulus of today when they remember how they once became people, in the days of Chaka, how the nations ate in the sun because of fear of them, even when they remember their fallen kingdom, they wince in their eyes, saying: "They're boiling, they're boiling! The springs are big!" &
        
         \vspace{.05cm}
         \texttt{HUM1}: So it came about, the end of Chaka, son of Senzangakhona. Even to this very day the Zulus, when they think how they were once a strong nation in the days of Chaka, and how other nations dreaded them so much that they could hardly swallow their food, and when they remember their kingdom which has fallen, tears well up in their eyes, and they say: “They ferment, they curdle! Even great pools dry away!” & 
        
         \vspace{.05cm}
         \texttt{HUM2}: And this was the last of Chaka, the son of Senzangakona. Even to-day the Mazulu remember how that they were men once, in the time of Chaka, and how the tribes in fear and trembling came to them for protection. And when they think of their lost empire the tears pour down their cheeks and they say: ‘Kingdoms wax and wane. Springs that once were mighty dry away.’
         \end{tabular}}
    \end{tabular}

    \caption{An example of one source paragraph in \name, from Nikolai Gogol's \textit{Dead Souls} (upper example) and from Thomas Mofolo's \textit{Chaka} (lower example) with their corresponding Google translation to English and aligned paragraphs from human-written translations.}
    \label{tab:par3_example}
\end{table*}

\subsection{Selecting works of literature}
For a source text to be included in \name, it must be (1) a literary work that has entered the public domain of its country of publication by 2022 with (2) a published electronic version along with (3) multiple versions of human-written, English translations. The first requirement skews our corpus towards older works of fiction. The second requirement ensures the preservation of the source texts' paragraph breaks. The third requirement limits us to texts that had achieved enough mainstream popularity to warrant (re)translations in English. Our most-recently published source text, \textit{The Book of Disquietude}, was published posthumously in 1982, 47 years after the author's death. The oldest source text in our dataset, \textit{Romance of the Three Kingdoms}, was written in the 14th-century. The full list of literary works with source language, author information, and publication year is available in Table \ref{tab:par3_all_data} in the Appendix.


\subsection{Translating works using Google Translate}
Before being fed to Google Translate, the data was preprocessed to convert ebooks to lists of plain text paragraphs and to remove tables of contexts, translator notes, and text-specific artifacts.\footnote{From Japanese texts, we removed artifacts of \textit{furigana}, a reading aid placed above difficult Japanese characters in order to help readers unfamiliar with higher-level ideograms.} Each paragraph was passed to the default model of the Google Translate API between April 20 and April 27, 2022. The total cost of source text translation was about 900 USD.\footnote{Google charges 20 USD per 1M characters of translation.}

\subsection{Aligning paragraphs}
All English translations, both human and Google Translate-generated, were separated into sentences using spaCy's Sentencizer.\footnote{\url{https://spacy.io/usage/linguistic-features\#sbd}} The sentences of each human translation were aligned to the sentences of the Google translation of the corresponding source text using the Needleman-Wunsch algorithm \citep{needleman-wunsch-1970} for global alignment. Since this algorithm requires scores between each pair of human-Google sentences, we compute scores using the  embedding-based \textsc{Sim} measure developed by~\citet{wieting-etal-2019-beyond}, which performs well on semantic textual similarity (STS) benchmarks~\citep{agirre-etal-2016-semeval}. Final paragraph-level alignments were computed using the paragraph segmentations in the original source text.


\subsection{Post-processing and filtering}
We considered alignments to be ``short'' if any English paragraph, human or Google generated, contained fewer than 4 tokens or 20 characters. We discarded any alignments that were ``short'' and contained the word ``chapter'' or a Roman numeral, as these were overwhelmingly chapter titles. We also discarded any alignments where one English paragraph contained more than 3 times the number of words than another, reasoning that these were actually misalignments. Thus, we also discarded any alignments with a \textsc{Bleu} score of less than 5. Alignments were sampled for the final version of \name\ such that no more than 50\% of the paragraphs for any human translation were included. Finally, alignments for each source text were then shuffled, at the paragraph level, to prevent reconstruction of the human translations, which may not be in the public domain.

\subsection{Train, test, and validation splits}
Instead of randomly creating splits of the 121K paragraphs in \name, we define  train, test, and validation splits at the document level. Each literary text belongs to one split, and all translations associated with its source paragraphs belong to that split as well. This decision allows us to better test the generalization ability of systems trained on \name, and avoid cases where an MT model memorizes entities or stylistic patterns located within a particular book to artificially inflate its evaluation scores.  The training split contains around 80\% of the total number of source paragraphs (97,611), the test split contains around 10\% (11,606), and the validation split contains around 10\% (11,606). Appendix~\ref{tab:par3_all_data} shows the texts belonging to each split.

\section{How good are existing MT systems for literary translation?}
\label{sec:analysis}
Armed with our \name\ dataset, we next turn to evaluating the ability of commercial-grade MT systems for literary translation. First, we describe a study in which we hired both professional literary translators and monolingual English experts to compare reference translations to those produced by Google Translate at a paragraph-level. In an A/B test, the translators showed a strong preference (on 84\% of examples) for human-written translations, finding MT output to be far too literal and riddled with discourse-level errors (e.g., pronoun consistency or contextual word sense issues). The monolingual raters preferred the human-written translations over the Google Translate outputs 85\% of the time, suggesting that discourse-level errors made by MT systems are prevalent and noticeable when the MT outputs are evaluated independently of the source texts. Finally, we address deficiencies in existing \emph{automatic} MT evaluation metrics, including \textsc{Bleu}, \textsc{Bleurt}, and the document-level \textsc{BloNDe} metric. These metrics failed to distinguish human from machine translation, even preferring the MT outputs on average. 

\subsection{Diagnosing literary MT with judgments from expert translators}
\label{subsec:human_eval}
As literary MT is understudied (especially at a document level), it is unclear how state-of-the-art MT systems perform on this task and what systematic errors they make. To shed light on this issue, we hire human experts (both monolingual English experts as well as literary translators fluent in both languages) to perform A/B tests on \name\ which indicates their preference of a Google Translate output paragraph (\texttt{GTr}) versus a reference translation written by a human (\texttt{HUM}). We additionally solicit detailed free-form comments for each example explaining the raters' justifications. We find that both monolingual raters and literary translators strongly prefer \texttt{HUM} over \texttt{GTr} paragraphs, noting that overly literal translation and discourse errors are the main error sources with \texttt{GTr}. 

\paragraph{Experimental setup:}
We administer A/B tests to two sets of raters: (1) monolingual English experts (e.g., creative writers or copy editors), and (2) professional literary translators. For the latter group, we first provided a source paragraph in German, French, or Russian. Under the source paragraph, we showed two English translations of the source paragraph: one produced by Google Translate and one from a published, human-written translation.\footnote{Each English paragraph was 130-180 words long.} We asked each rater to choose the ``better'' translation and also to give written justification for their choice (2-3 sentences). While all raters knew that the texts were translations, they did \textbf{NOT} know that one paragraph was machine-generated. Each translator completed 50 tasks in their language of expertise. For the monolingual task, the set up was similar except for two important distinctions: (1) \textbf{NO} source paragraph was provided and (2) each monolingual rater rated all 150 examples (50 from each of 3 language-specific tasks). 
Tasks were designed and administered via Label Studio,\footnote{\url{https://labelstud.io/}} an open-source data-labeling tool, and raters\footnote{For the language-specific task, raters were required to be professional literary translators with experience translating German, French, or Russian to English. We hired one translator for each language. For the monolingual task, we hired three raters with extensive experience in creative writing, copy-editing, or English literature.} were hired using Upwork, an online platform for freelancers.\footnote{\url{https://www.upwork.com/}} For the completion of 50 language-specific tasks, translators were paid \$200 each. For the set of 150 monolingual tasks, raters were paid \$250 each. All raters were given at least 4 days to complete their tasks.

\paragraph{Common MT errors:} We roughly categorize the errors highlighted by the professional literary translators into five groups. The most pervasive error (constituting nearly half of all translation errors identified) is the \textbf{overly literal} translation of the source text, where a translator adheres too closely to the syntax of the source language, resulting in awkward phrasing or the  mistranslation of idioms. The second most prevalent errors are \textbf{discourse} errors, such as pronoun inconsistency or coreference issues, which occur when context is ignored--these errors are exacerbated at the paragraph and document levels. We define the rest of the categories and report their the distribution in Table \ref{tab:common_mt_errors}.

\begin{table*}[h]
\tiny
\centering
    \begin{tabular}{ p{10.2cm}|p{1.5cm}|p{3cm}} 
        \toprule
        \bf Example & \textbf{Error Type (\%)}& \bf Translator Comments \\
        \midrule
        From \textit{The Sin of Abbé Mouret}, Emile Zola\newline\newline
        \texttt{SRC}: L’abbé Mouret dépensa \violet{là} ses économies du séminaire. C’étaient, d’ailleurs, des embellissements dont la naïveté maladroite eût fait sourire. La maçonnerie le rebuta vite. Il se contenta de recrépir le tour de l’église, à hauteur d’homme. La Teuse \blue{gâchait} le plâtre.\newline \newline \texttt{HUM}: Abbé Mouret spent all his seminary savings \violet{on the work}. His embellishments were so clumsy and naive as to raise a smile. The masonry-work soon lost its appeal for him. He contented himself with replastering all round the church to the height of a man’s head. La Teuse \blue{mixed} the plaster.\newline & \violet{Discourse}\newline(20.8\%)\newline Issues created by lack of context. & The first line in French includes the adverb \violet{"là" which means "there"}. In the translation I selected, \violet{"là" is translated by "on the work"} to mean that that is what the Abbé spent all his money on. It makes it easier to understand and the text flows better.\newline\\
        \texttt{GTr}: Father Mouret spent his seminary savings \violet{there}. They were, moreover, embellishments whose clumsy simplicity would have made you smile. Masonry soon put him off. He contented himself with replastering around the church, at eye level. La Teuse \blue{ruined} the plaster.&  \blue{Word sense}\newline(7.3\%)\newline Incorrect translation chosen where multiple are valid. & The verb \blue{"gâcher" usually means "to waste" / "to ruin"}. However, when used with "plâtre" (=plaster), it means \blue{"to mix" / "to temper"}--This is a collocation that the author of the second translation missed \dots but that was translated correctly in the passage I selected.\\
        \midrule
        From \textit{We}, Yevgeny Zamyatin\newline\newline
        \texttt{SRC}: \begin{otherlanguage}{russian}Проснулся: умеренный, синеватый свет; блестит стекло стен, стеклянные кресла, стол. \teal{ Это успокоило, сердце перестало колотиться}. \red{Сок}, Будда… что за абсурд? Ясно: болен. Раньше я никогда не видел снов. Говорят, у древних это было самое обыкновенное и нормальное – видеть сны. Ну да: ведь и вся жизнь у них была вот такая ужасная карусель: зеленое – оранжевое – Будда – \red{сок}. Но мы-то знаем, что сны – это серьезная психическая болезнь. И я знаю: до сих пор мой мозг был хронометрически выверенным, сверкающим, без единой соринки механизмом, а теперь… Да, теперь именно так: я чувствую там, в мозгу, какое-то инородное тело – как тончайший ресничный волосок в глазу: \orange{ всего себя чувствуешь, а вот этот глаз с волоском – нельзя о нем забыть ни на секунду…}\end{otherlanguage} \newline & \orange{Overly literal}\newline(48.4\%)\newline The translation adheres too closely to the syntax of the source language.& The \orange{last sentence} of the passage is pretty tough and requires an understanding of the context. The author of the first translation did a great job and conveyed the meaning of the source sentence properly. The author of the second translation made a mistake by using \orange{word-by-word translation: "everything you feel yourself."} As a result, the phrase makes no sense.\\
        \texttt{HUM}: I woke: soft, bluish light, glimmer of glass walls, glass chairs and table. \teal{This calmed me; my heart stopped hammering.} \red{Sap}, Buddha ... what nonsense! Clearly I must be ill. I have never dreamed before. They say that with the ancients dreaming was a perfectly ordinary, normal occurrence. But of course, their whole life was a dreadful whirling carousel---green, orange, Buddhas, \red{sap}. We, however, know that dreams are a serious psychic disease. And I know that until this moment my brain has been a chronometrically exact gleaming mechanism without a single speck of dust. But now \dots Yes, precisely: I feel some alien body in my brain, like the finest eyelash in the eye. \orange{You do not feel your body, but that eye with the lash in it---you can't forget it for a second.}\newline & \red{Precision}\newline(7.3\%)\newline The translation is either too specific or not specific enough. &The author of the source text mentions "\begin{otherlanguage}{russian}\red{сок}\end{otherlanguage}" which can be translated as "\red{sap}" (as in the first translation). The author of the second translation decided to transcribe this word in one sentence as "\red{Sok}" (which doesn't convey the meaning of the Russian word at all) and then translated it as "\red{juice}". \newline\\
        \texttt{GTr}: Awake: moderate, bluish light; glittering glass walls, glass chairs, table. \teal{It calmed her down and her heart stopped beating.} \red{Sok}, Buddha... what an absurdity? Obviously sick. I have never dreamed before. They say that among the ancients it was the most ordinary and normal thing---to dream. Well, yes: after all, their whole life was such a terrible carousel: green - orange - Buddha - \red{juice}. But we know that dreams are a serious mental illness. And I know: until now, my brain was a chronometrically verified, sparkling, without a single mote mechanism, but now\dots Yes, now it’s exactly like this: I feel there, in the brain, some kind of foreign body---like the thinnest ciliary hair in the eye: \orange{everything you feel yourself, but this eye with a hair---you can’t forget about it for a second\dots} &\teal{Catastrophic}\newline(16.1)\%\newline Errors that completely invalidate the translation. & According to the source text the narrator is male and he tells a story about himself. There is \teal{a sentence "It calmed her down and her heart stopped beating" in the second translation which makes no sense} if we compare it to the Russian text. \\
        \bottomrule
    \end{tabular}
    \caption{Definitions and examples of the five types of translation errors on Google Translate outputs identified by professional literary translators. We report their prevalence as a percentage of all errors identified by the translators and include the translators' explanations.}
    \label{tab:common_mt_errors}
\end{table*}

\begin{figure}[h!]
\centering
\includegraphics[width=.9 \columnwidth]{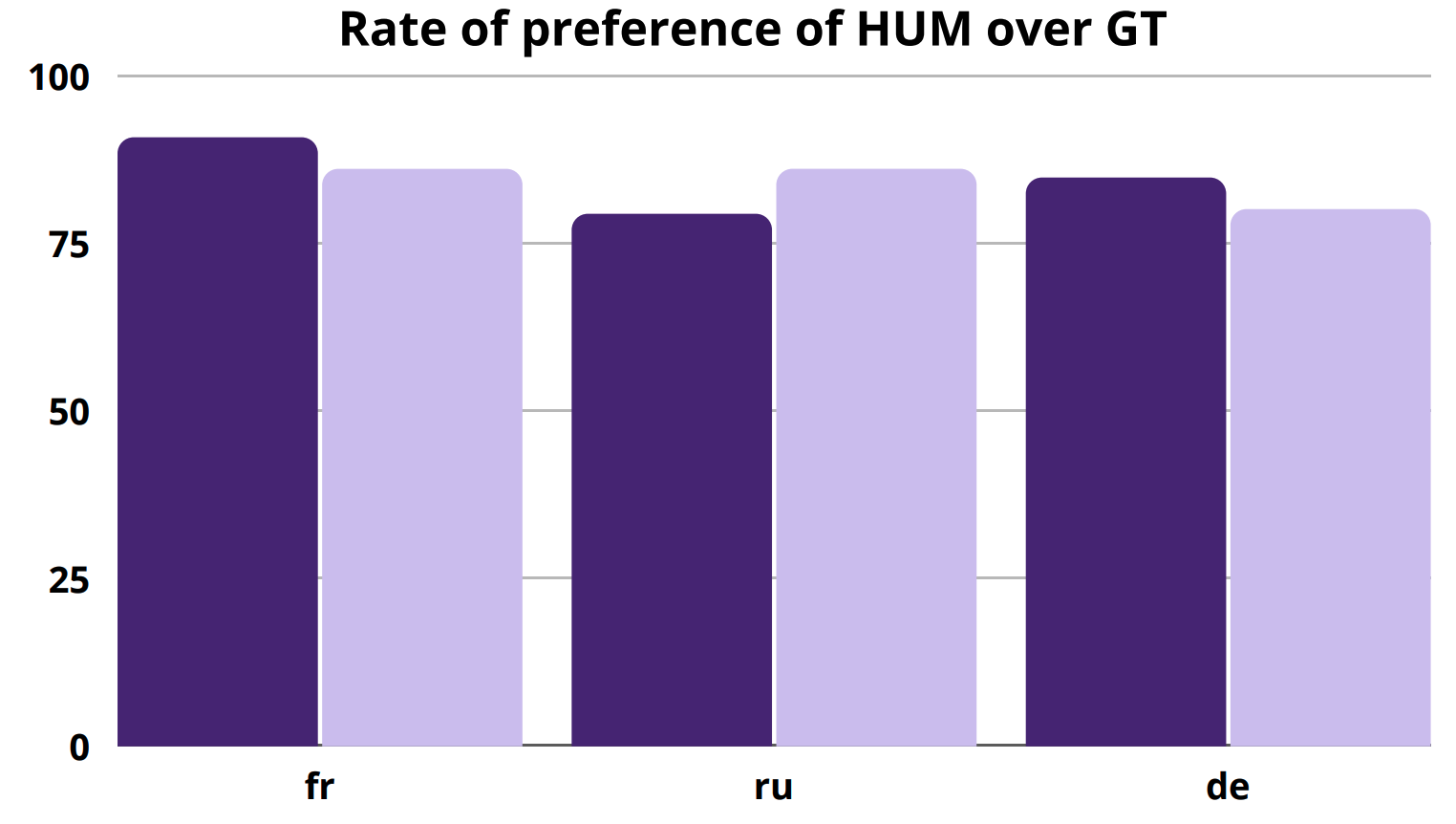}
\caption{The percentage of cases in which raters preferred the human-written translation to the Google translation by source language. Note that the value for monolingual raters is the average of 3 percentages for 3 monolingual raters.}
\label{figure:hum_vs_GTr}
\end{figure}

\paragraph{Monolingual vs translator ratings:}
Though the source text is essential to the practice of translation, the monolingual setting of our A/B testing allows us to identify attributes other than translation errors that distinguish the MT system outputs from human-written text. Both monolingual and bilingual raters strongly preferred \texttt{HUM} to \texttt{GTr} across all three tested languages\footnote{We report Krippendorff`s alpha \citep{Krippendorff2011ComputingKA} as the measure of inter-annotator agreement (IAA). The IAA between the monolingual raters was 0.546 (0.437 for Russian,  0.494 for German, and 0.707 for French). The IAA between the aggregated votes of monolingual raters (majority vote) and the translator was 0.524 for Russian, 0.683 
for German, and 0.681 for French. These numbers suggest moderate to substantial agreement \citep{artstein-poesio-2008-survey}.}, as shown in Figure \ref{figure:hum_vs_GTr}, although their preference fell on Russian examples.
In a case where all 3 monolingual raters chose \texttt{HUM} while the translator chose \texttt{GTr}, their comments reveal that the monolingual raters prioritized clarity and readability:
\begin{quote}
    \footnotesize
    [\texttt{HUM}] ``is preferable because it flows better and makes better sense'' and ``made complete sense and was much easier to read''
\end{quote}
while the translator diagnosed \texttt{HUM} with a catastrophic error: 
\begin{quote}
    \footnotesize
    ``[\texttt{HUM}] contains several mistakes, mainly small omissions that change the meaning of the sentence, but also wrong translations (`trained European chef' instead of `European-educated chef').''
\end{quote}

For an example where all 3 monolingual raters chose [\texttt{GTr}] while the translator chose [\texttt{HUM}], the monolingual raters much preferred the contemporary language in [\texttt{GTr}]:
\begin{quote}
    \footnotesize
    [\texttt{GTr}] was ``much easier for me to grasp because of its structure compared to the similar sentence in [\texttt{HUM}]'' and praised for its ``use of commonplace vocabulary that is understandable to the reader.'' 
\end{quote}

However, the translator, with access to the source text, identified a precision error in \texttt{GTr}, and ultimately declared \texttt{HUM} to be the better translation: 
\begin{quote}
    \footnotesize
    ``\textit{lord} from [\texttt{HUM}] is the exact translation of the Russian \begin{otherlanguage}{russian}бари \end{otherlanguage}  while \textit{bard} from [\texttt{GTr}] doesn't convey a necessary meaning.''\footnote{To view the \texttt{SRC}, \texttt{HUM}, and \texttt{GTr} texts for these examples, see Tables \ref{tab:hum_v_gt_1} and \ref{tab:hum_v_gt_2} in the Appendix.}
\end{quote}

\subsection{Can automatic MT metrics evaluate literary translation?}

\begin{table}[ht]
    \renewcommand{\arraystretch}{1.15}
    \centering
    \scriptsize
    \begin{tabular}{c|cc|cc|cc}
        \hline
        \multirow{2}{*}{\begin{tabular}[c]{@{}c@{}}Source\\ lang\end{tabular}} & \multicolumn{2}{c|}{\textsc{Bleu}} & \multicolumn{2}{c|}{\textsc{Bleurt}} & \multicolumn{2}{c}{\textsc{BlonDe}} \\ \cline{2-7} 
         & \multicolumn{1}{c|}{\texttt{HUM}} & \texttt{GTr} & \multicolumn{1}{c|}{\texttt{HUM}} & \texttt{GTr} & \multicolumn{1}{c|}{\texttt{HUM}} & \texttt{GTr} \\ \hline
         \textit{fr} & 26.8 & 29.4 & 0.630 & 0.630 & 25.6 & 27.5 \\ 
         \textit{ru} & 28.8 & 29.6 & 0.642 & 0.622 & 25.2 & 26.0 \\ 
         \textit{de} & 23.1 & 24.6 & 0.598 & 0.597 & 22.0 & 23.6 \\ 
         \textit{no} & 29.0 & 26.5 & 0.628 & 0.595 & 28.3 & 29.6 \\
         \textit{es} & 24.8 & 22.4 & 0.623 & 0.547 & 27.4 & 24.2 \\  
         \textit{cs} & 15.4 & 20.4 & 0.560 & 0.566 & 14.6 & 20.2 \\
         \textit{sv} & 36.7 & 36.4 & 0.680 & 0.669 & 39.5 & 41.0 \\ 
         \textit{pt} & 31.8 & 27.9 & 0.646 & 0.598 & 29.2 & 27.3 \\ 
         \textit{it} & 21.8 & 24.6 & 0.646 & 0.628 & 23.3 & 24.8 \\ 
         \textit{ja} & 14.8 & 12.5 & 0.568 & 0.512 & 15.0 & 12.5 \\
         \textit{bn} & 10.4 & 12.1 & 0.596 & 0.572 & 9.9 & 11.1 \\ 
         \textit{ta} & 15.5 & 14.6 & 0.581 & 0.561 & 11.2 & 10.4 \\
         \textit{da} & 26.7 & 25.5 & 0.614 & 0.566 & 19.1 & 16.8 \\ 
         \textit{zh} & 11.8 & 11.7 & 0.482 & 0.434 & 8.7 & 8.8 \\
         \textit{nl} & 26.0 & 23.9 & 0.640 & 0.625 & 23.1 & 22.3 \\ 
         \textit{hu} & 26.3 & 19.5 & 0.640 & 0.602 & 26.4 & 18.7 \\ 
         \textit{pl} & 34.89 & 18.5 & 0.667 & 0.563 & 28.2 & 14.8 \\
         \textit{st} & 16.9 & 15.78 & 0.559 & 0.499 & 16.4 & 14.7 \\
         \textit{fa} & 15.2 & 16.2 & 0.540 & 0.503 & 8.9 & 11.0 \\ \hline 
         \textbf{All} & 26.4 & 27.6 & 0.536 & 0.613 & 24.5 & 25.6 \\
         Win \% & 38.2\% & 61.8\% & 53.6\% & 46.4\% & 37.4\% & 62.6\%\\
         \hline
    \end{tabular}
    \caption{Average \textsc{Bleu}, \textsc{Bleurt}, and \textsc{BlonDe} scores for \name\ by source language, computed using the same reference set on human and Google translations. See Section 3.2 for details on the computation of the average metric score. The Win \% in the final row is the percentage of cases, out of 121,385 unique source paragraphs, in which the metric prefers the human or the Google translation.}
    \label{tab:par3_auto_metrics}
\end{table}

    

Expert human evaluation, while insightful, is also time-consuming and expensive, which precludes its use in most model development scenarios. The MT community thus relies extensively on \emph{automatic} metrics that score candidate translations against references. In this section, we explore the usage of three metrics (\textsc{Bleu}, \textsc{Bleurt}, and \textsc{BlonDe}) on literary MT evaluation, and we discover that none of them can accurately distinguish \texttt{GTr} text from \texttt{HUM}. Regardless of their performance, we also note that most automatic metrics are designed to work with sentence-level alignments, which are rarely available for literary translations because translators merge and combine sentences. Thus, developing domain-specific evaluation metrics is crucial to make meaningful progress in literary MT.

\paragraph{MT Metrics:}
To study the ability of MT metrics to distinguish between machine and human translations, we compute three metrics on \name:

    \noindent\textbf{\textsc{Bleu}}~\citep{bleu}\footnote{We compute the default, case-sensitive implementation of \textsc{Bleu} from \url{https://github.com/mjpost/sacrebleu}.} is a string-based multi-reference metric originally proposed to evaluate sentence-level translations but also used for document-level MT~\citep{liu2020multilingual}.
    
    \noindent\textbf{\textsc{Bleurt}}~\citep{sellam-etal-2020-bleurt} is a pretrained language model fine-tuned on human judgments of translation-reference pairs.\footnote{We compute \textsc{Bleurt} for \name\ using the recommended and most recent checkpoint, \textsc{Bleurt-20}. As the maximum input length for \textsc{Bleurt} is 512 sentencepiece tokens, we exclude inputs which exceed this length and would be otherwise truncated. In total, 1.4\% of the dataset was excluded.} \textsc{Bleurt} has been shown to be effective on document-level tasks such as summarization~\citep{kasai2021bidimensional}.
    
    \noindent\textbf{\textsc{BlonDe}}~\citep{jiang-etal-2022-blonde}\footnote{We compute \textsc{BlonDe.F1}, simply referred to as \textsc{BlonDe} in the original paper.} is a document-level multi-reference evaluation metric that considers discourse coherence by calculating the F1 of four ``discourse categories'' that each represent a feature of coherence across sentences, such as \textbf{tense} or \textbf{pronoun} consistency. 

\begin{figure*}[t!]
\centering
\includegraphics[width=0.98\linewidth]{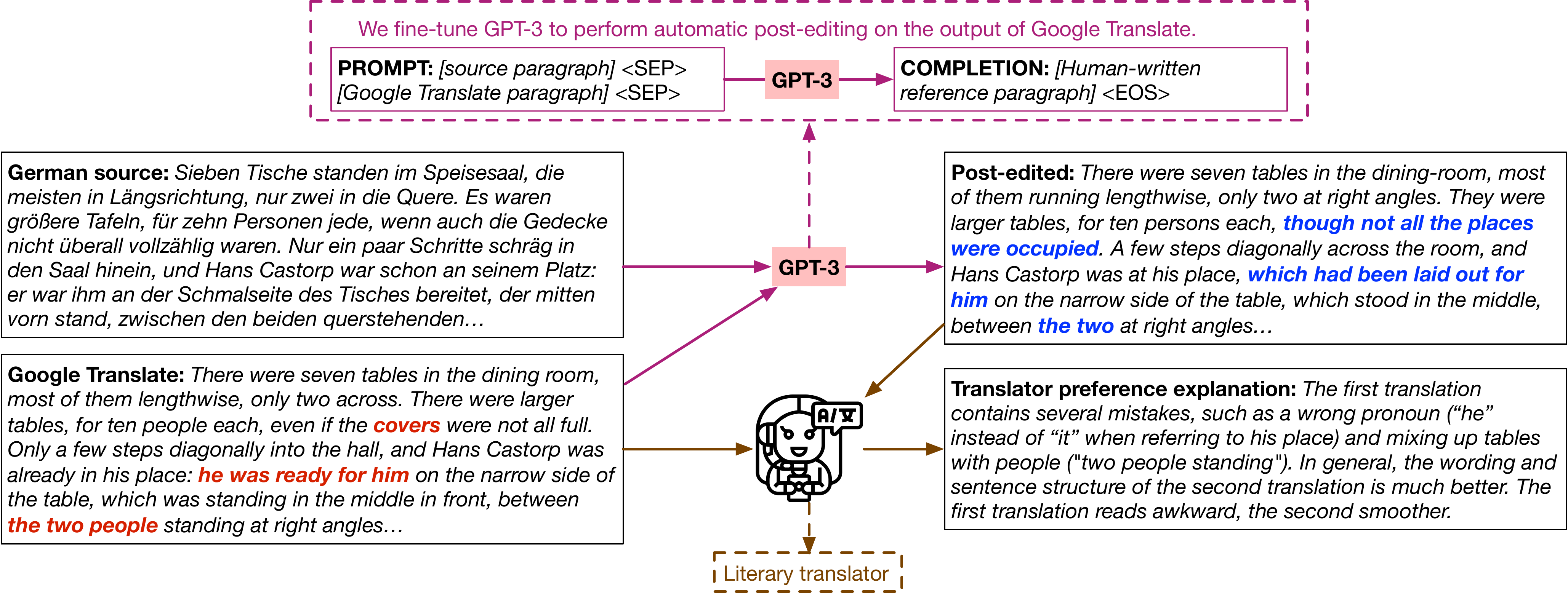}
\caption{An illustration of our automatic post-editing model on a \name\ source paragraph from Thomas Mann's \emph{The Magic Mountain}, which fine-tunes GPT-3 to transform a Google Translate paragraph into a human-written reference translation. We hire professional literary translators (in this case, a German translator) to perform a blind A/B test comparing Google Translate against the post-edited output and also to provide free-form explanations as to why they made their choice. In this case, and \textbf{69.3\%} of the time overall, they prefer the post-editing model’s output.}
\label{figure:postedit}
\vspace{-2mm}
\end{figure*}

\paragraph{Comparing \texttt{HUM} to \texttt{GTr}:} Since \name\ contains a variable number of references for each paragraph, we aggregate metric scores across all references for fair comparison, following the methodology of prior work in crowdsourcing multiple reference translations~\citep{callison-burch-2009-fast,zaidan-callison-burch-2011-crowdsourcing}. Given a specific source paragraph with an aligned translation $\texttt{GTr}$ produced by Google Translate and a set of $n$ human reference paragraphs $\texttt{HUM}_{1\dots n}$, we compute aggregate  scores ($s_{{HUM}}$) of a given metric \textsc{Metric} for human references (against each other) as:

\begin{equation*}
    s_{HUM} = \sum_i \frac{\textsc{Metric}(\texttt{HUM}_i, \texttt{HUM}_{1\dots n} - \{\texttt{HUM}_i\})}{n}
\end{equation*}

We use the same reference sets for each example to compute aggregate scores for Google Translate outputs, which ensures that the numbers are comparable:

\begin{equation*}
    s_{GTr} = \sum_i \frac{\textsc{Metric}(\texttt{GTr}, \texttt{HUM}_{1\dots n} - \{\texttt{HUM}_i\})}{n}
\end{equation*}

For \textsc{Bleurt}, which unlike \textsc{Bleu} and \textsc{BlonDe} is not well-defined for multiple references, we compute \textsc{Bleurt}$(\texttt{GTr}, \texttt{HUM}_{1\dots n} - \{\texttt{HUM}_i\})$ by taking the average over pairwise \textsc{Bleurt} scores between \texttt{GTr} and each reference.

\paragraph{Automatic metrics are not predictive of literary MT quality:}  We have identified mistakes made by Google Translate in Section~\ref{subsec:human_eval} and the human translations have all been professionally edited and published. Hence, we expect automatic metrics to prefer the human translations. However, we show in Table \ref{tab:par3_auto_metrics} that two of our three metrics, \textsc{Bleu} and \textsc{BlonDe},\footnote{\citet{jiang-etal-2022-blonde} show that \textsc{BlonDe} has very high correlation to \textsc{Bleu}.} fail to distinguish meaningfully between the human and Google translations, preferring the Google translation to the human one in over 60\% of cases. For \textsc{Bleurt}, the choice between Google and human is nearly chance, with human translations preferred 53.6\% of the time. 
A Wilcoxon signed-rank test \citep{Wilcoxon1945} reveals that both \textsc{Bleu} (\textit{z}=-67.344, \textit{p}<.001, \textit{r}=.192) and \textsc{BlonDe} (\textit{z}=-62.862, \textit{p}<.001, \textit{r}=.179) prefer Google Translate over human translation. \textsc{Bleurt}, on the other hand, appears to correctly distinguish between the human translation and Google Translate (\textit{z}=42.462, \textit{p}<.001, \textit{r}=.118); however, the effect size is small (\textit{r}<.30) in all three cases.\footnote{We also perform bootstrapping which yields comparable results.}

\section{Can automatic post-editing improve literary MT?}
\label{sec:postediting}

From the experiments in the previous section, we can conclude that human expert evaluation is currently the only way to judge the quality of literary MT. We now turn to improving the quality of Google Translate outputs on \name\ via \emph{automatic} post-editing~\citep{chatterjee2018findings}, in which a model corrects the output of a black-box MT system. While~\citet{toral2018post} show that manual post-editing on top of MT outputs aids human translator efficiency in the literary domain, no prior work has applied automatic post-editing to literary translations. As shown in Figure \ref{figure:postedit}, we feed both the source paragraph and the Google Translate output to the GPT-3~\citep{brown2020language} language model, which has been shown to have zero-shot translation capability (although far below state-of-the-art supervised MT systems). We fine-tune GPT-3 to produce a human-written reference translation given these inputs and find that it mitigates issues with overly literal translation and discourse errors.

\subsection{Literary post-editing with GPT-3}

Our analysis experiments reveal that discourse-level errors that span multiple sentences (e.g., coreference, stylistic consistency, contextual word sense selection) are a huge problem for Google Translate when applied to literary MT. Motivated to address these issues, we select the 175B parameter GPT-3 \emph{davinci} model as our base post-editing system, as it can operate over paragraph-length inputs (max sequence length of 2048 tokens), encode text in multiple languages, and exhibits impressive ability to learn complex tasks with limited training data. To form fine-tuning examples for GPT-3, we concatenate a source paragraph \texttt{SRC}, an aligned Google Translate paragraph \texttt{GTr}, and a human reference translation \texttt{HUM} using special separator and end-of-sequence tokens:\footnote{For the separator token between the source and the Google translation paragraphs, we arbitrarily chose "\#\#" and for the separator token between the prompt and the completion, we used "\textbackslash n\textbackslash n\#\#\#\textbackslash n\textbackslash n" as recommended by OpenAI guidelines. The EOS token (stop sequence) was "DNE".}

\begin{quote}
    seq = \texttt{SRC} \texttt{<SEP>} \texttt{GTr} \texttt{<SEP>} \texttt{HUM} \texttt{<EOS>}
\end{quote}

where \texttt{SRC <SEP> GTr} is considered the \textit{prompt} and \texttt{HUM <EOS>} is the \textit{completion}. 

\paragraph{Data filtering:} 
Before fine-tuning our model, we filtered the \name\ training set to remove examples where the aggregated \textsc{Bleu} scores between \texttt{GTr} and \texttt{HUM} were either in the 10$^{th}$ or 90$^{th}$ percentiles, which ignores both noisy alignments and near-perfect \texttt{GTr} outputs that do not need any edits. For each example, we also only use the \texttt{HUM} paragraph with the maximum \textsc{Bleu} against the \texttt{GTr} output for that source paragraph, since we could not use all references during fine-tuning.\footnote{We excluded any examples in which the total number of tokens in the source, \texttt{GTr}, and human paragraphs was greater than 2,000, as GPT-3 training examples (prompt \textit{and} completion) must be fewer than 2,048 tokens.} 
 Finally, we randomly sample 30K of the filtered training examples because of the prohibitive cost of fine-tuning and using the GPT-3 \emph{davinci} model.\footnote{\emph{davinci} costs 0.03 USD per 1k tokens to fine-tune and 0.12 USD per 1k tokens to use a fine-tuned model.} See Appendix~\ref{sec:config} for our fine-tuning configuration.


\subsection{Human evaluation of post-edited outputs} 
Having established that human evaluation is critical for literary MT, we had the same 3 professional translators perform A/B testing on \texttt{GTr} and the outputs of our post-editing model \texttt{GPT-3}.\footnote{We report the scores of the 3 automatic MT metrics on the outputs of GPT-3 in Table \ref{tab:par3_test_metrics} in the Appendix.} The translators prefer \texttt{GPT-3} over \texttt{GTr} at a rate of 69\% (\textit{p}<.001, 95\% CI [0.613, 0.770]). The comments show that the model often improved on the ``overly literal'' nature of many \texttt{GTr} paragraphs: 
\begin{quote}
    \footnotesize
    ``The phrase \begin{otherlanguage}{russian}с знакомыми, очень знакомыми улыбкой и взглядом \end{otherlanguage} from the source text should be translated as `with a familiar, very familiar smile and gaze' as in [\texttt{GPT-3}]. The author of [\texttt{GTr}] makes mistakes in choosing the words and suggests "with acquaintances, very familiar smile and look.''\footnote{See Table \ref{tab:gpt3_v_gt} in the Appendix for the texts.}
\end{quote}

\begin{figure}[h!]
\centering
\includegraphics[width=.9\columnwidth]{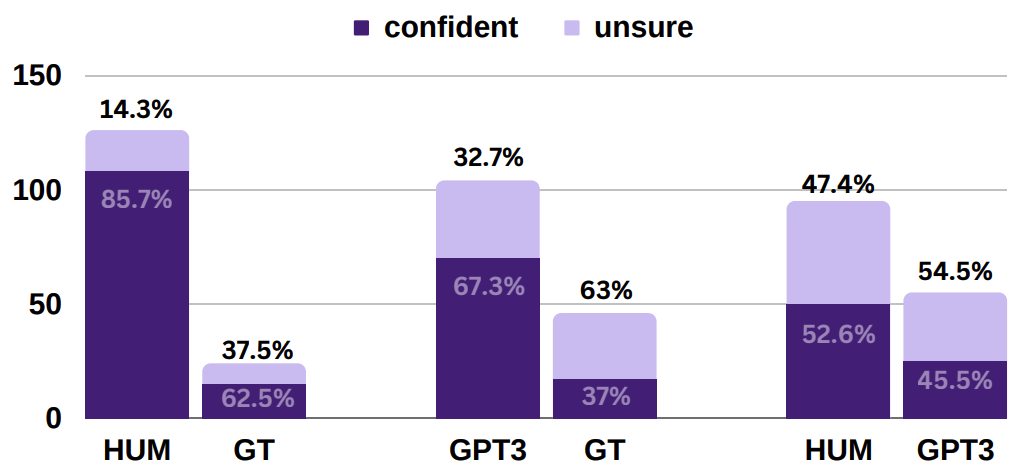}
\caption{The number of votes for \texttt{HUM} vs \texttt{GTr},  \texttt{GPT-3} vs \texttt{GTr}, and  \texttt{HUM} vs \texttt{GPT-3} along with their corresponding raters' confidence.}
\label{figure:annotator_confidence}
\end{figure}

Finally, we also had the 3 professional translators perform A/B testing on the post-edited \texttt{GPT-3} outputs and \texttt{HUM}. While the translators still preferred \texttt{HUM}, their preference rate decreased from 84\% (vs. \texttt{GTr}) to 63\%.  Their comments reveal an interesting caveat to their judgments: overall, raters are much more confident when selecting \texttt{GPT-3} than when selecting \texttt{GTr} when choosing between the two machine translations. When they did choose \texttt{GTr}, they were often unsure because both translations were equivalently good or bad. When comparing \texttt{HUM} to \texttt{GPT-3}, our annotators were unsure around half of the time, regardless of whether they selected \texttt{HUM} or \texttt{GPT-3} (they were slightly more confident when choosing \texttt{HUM}), suggesting that the task of discerning the better translator was particularly challenging. We present the results of a small-scale quantitative analysis of the 150 comments across the 3 raters in Figure \ref{figure:annotator_confidence}. 

\paragraph{Characterizing the behavior of \texttt{GPT-3} post-editing:} We performed a fine-grained analysis of the comments provided by professional translators regarding the behavior of the \texttt{GPT-3} post-editing model. Overall, the translators observe several positives, including correcting pronoun errors and mistranslations in addition to better capturing the sense of the original work compared to \texttt{GTr}. For example, the professional Russian translator noted an instance where \texttt{GPT-3} resolved \texttt{GTr}'s inconsistent use of character names:
\begin{quote}
    \footnotesize
    The narrator mentions one character whose name is “Ippolit.” The author of the [\texttt{Gtr}] translation uses this spelling, but then changes it into ``Hippolyte'' for no reason.
\end{quote}
On the other hand, the \texttt{GPT-3} text occasionally omits some details or contains stylistic choices with which the translators disagree:
\begin{quote}
    \footnotesize
    The only reason I selected [\texttt{GTr}] is because I like how it kept the full French nickname ``La Teuse'' and did not translate the determiner into ``the Teuse'' [as \texttt{GPT-3} did].
\end{quote}

We show more examples of the post-edit model correcting errors in \texttt{GTr} in Table \ref{tab:post_edit_examples}. In 31 out of 150 cases, the translators felt that \texttt{GPT-3} did not notably improve upon \texttt{GTr}, often mentioning that neither translation was preferred.

\paragraph{Did \texttt{GPT-3} see the
translations in pretraining?}
One potential criticism of the \texttt{GPT-3} post-editing model is that it may have seen the reference translations in its pretraining data, and thus any improvements could just be a result of memorization rather than actual understanding of the source text. We qualitatively measure this by creating a small dataset of translated paragraphs that could not have been seen by \texttt{GPT-3}. These translations were either published after \texttt{GPT-3} was pretrained (2022), or manually translated by an author of this paper from previously untranslated works of literature. Even on this previously unseen data, our model can correct mistranslations, grammatical errors, and stylistic inconsistencies:
\begin{CJK*}{UTF8}{gbsn}
    \begin{quote}
        \footnotesize
        \texttt{SRC: }朱丽默默走下楼去，都没坐电梯，一路回想惊鸿一瞥的明成的脸。\\
        
        \texttt{GTr: }Zhu Li walked downstairs silently, without taking the elevator, all the way back to the face of Ming Cheng who had caught a glimpse.\\
        
        \texttt{GPT-3: }Zhu Li walked downstairs in silence, without taking the elevator, and all the way back she kept recalling the face of Ming Cheng, which she had seen for a moment.
    \end{quote}
\end{CJK*}

\section{Related Work}
Our work builds on previous work in literary machine translation. Some early work focused on \emph{poetry} translation~\citep{genzel-etal-2010-poetic,jones-irvine-2013-un}, which has recently been tackled with neural approaches ~\citep{chakrabarty-etal-2021-dont}. Other works have targeted novels, like those in \name, with focuses on manual post-editing~\citep{toral2018post,toral2020machine} and comparisons of neural MT to statistical MT systems~\citep{moorkens2018translators,toral2018level,Toral2015TranslatingLT}. Most of these works experiment with datasets far smaller than \name ~\citep{arenas2022creamt, fonteyne_literary_2020}. More recent work has involved studying the linguistic characteristics of post-edited literary machine-translated text \citep{castilho-2022, macken-2022-pe}.

Work towards document-level NMT has built on sentence-level MT ~\citep{tiedemann-scherrer-2017-neural,Jean2017DoesNM,bawden-etal-2018-evaluating,miculicich-etal-2018-document,agrawal_contextual_2018}. The a critical lack of parallel document-level corpora has inspired the creative use of parallel sentence-level data ~\citep{zhang-etal-2018-improving} and techniques for creating parallel document-level data ~\citep{junczys-wmt}. Our work also builds on efforts to tackle discourse-level errors specific to document-level MT and is very similar to that of ~\citet{voita-etal-2019-context}, but we specifically focus on the literary domain.

\section{Conclusion}
\label{sec:conclusion}
We study document-level literary machine translation by collecting a dataset (\name) of 121K parallel paragraphs from 104 novels. Our experiments show that existing automatic metrics of translation quality are not meaningful in the literary domain. A human evaluation experiment with professional literary translators reveals that commercial-grade MT systems are too literal in their translations and also suffer from discourse-level errors. We mitigate these problems to a certain extent by developing an automatic post-editing model using GPT-3. Overall, our work uncovers new challenges to progress in literary MT, and we hope that the public release of \name\ will encourage researchers to tackle them. 

\section*{Limitations}
\label{sec:limitations}
While \name\ covers a diverse array of genres and languages, there are potential confounding factors in the translation data to be aware of when performing analysis or modeling on top of it. First, multiple human translations of the same source text may not have been written independently: a later translator might have used an earlier translation as a reference, or a new translation may be commissioned because of dissatisfaction with older translations. Additionally, translators in our dataset differ in aspects such as years of experience, familiarity with the author of the source text (some were the exclusive translator for a single author), and bilinguality. The circumstances of each translation are also unique geographically and temporally. It is unclear whether (or how) to model such differences computationally, but it is an intriguing direction for future work.

We also acknowledge that our dataset has a single target-language; the curation of data in other target languages and the improvement of literary MT for other target languages is an essential step towards an equitable and more culturally-conscious field of NLP.

\section*{Ethical Considerations}

We acknowledge that the vast majority of the authors of our source texts are male. Because literary translation requires training, time, and money, our source texts skew towards older texts that achieved international popularity. We hope that our efforts towards better literary MT can aid literary translators in sharing more minority voices. The experiments involving humans were IRB-approved, and each hired rater was fairly compensated, with wages adjusted as we determined the average amount of time each task took.  
\section*{Acknowledgements}

We would like to thank the translators and English language professionals hired on Upwork for the efforts they put in the evaluation and their insightful comments. We would also like to show our appreciate to Tu Vu for sharing his knowledge about MT evaluation metrics and to Sergiusz Rzepkowski for his help in cleaning the data, as well as multiple translators whom we consulted when exploring our dataset. Finally, we would like to thank the UMass NLP community for their insights and discussions during this project.
This project was partially supported by awards IIS-1955567 and IIS-2046248 from the National Science Foundation (NSF) as well as an award from Open Philanthropy.

\bibliography{bib/anthology, bib/custom}

\begin{thebibliography}{48}
\expandafter\ifx\csname natexlab\endcsname\relax\def\natexlab#1{#1}\fi

\bibitem[{Agirre et~al.(2016)Agirre, Banea, Cer, Diab, Gonzalez-Agirre,
  Mihalcea, Rigau, and Wiebe}]{agirre-etal-2016-semeval}
Eneko Agirre, Carmen Banea, Daniel Cer, Mona Diab, Aitor Gonzalez-Agirre, Rada
  Mihalcea, German Rigau, and Janyce Wiebe. 2016.
\newblock \href {https://doi.org/10.18653/v1/S16-1081} {{S}em{E}val-2016 task
  1: Semantic textual similarity, monolingual and cross-lingual evaluation}.
\newblock In \emph{Proceedings of the 10th International Workshop on Semantic
  Evaluation ({S}em{E}val-2016)}, pages 497--511, San Diego, California.
  Association for Computational Linguistics.

\bibitem[{Agrawal et~al.(2018)Agrawal, Turchi, and
  Negri}]{agrawal_contextual_2018}
Ruchit Agrawal, Marco Turchi, and Matteo Negri. 2018.
\newblock Contextual {Handling} in {Neural} {Machine} {Translation}: {Look}
  {Behind}, {Ahead} and on {Both} {Sides}.

\bibitem[{Arenas and Toral(2022)}]{arenas2022creamt}
Ana~Guerberof Arenas and Antonio Toral. 2022.
\newblock Creamt: Creativity and narrative engagement of literary texts
  translated by translators and nmt.
\newblock In \emph{Proceedings of the 23rd Annual Conference of the European
  Association for Machine Translation}, pages 355--356.

\bibitem[{Artstein and Poesio(2008)}]{artstein-poesio-2008-survey}
Ron Artstein and Massimo Poesio. 2008.
\newblock \href {https://doi.org/10.1162/coli.07-034-R2} {Survey article:
  Inter-coder agreement for computational linguistics}.
\newblock \emph{Computational Linguistics}, 34(4):555--596.

\bibitem[{Baker(2018)}]{Baker2018-cg}
Mona Baker. 2018.
\newblock \emph{In other words}, 3 edition.
\newblock Routledge, London, England.

\bibitem[{Baker and Saldanha(2021)}]{Baker2021-qr-encyclopedia-transstudy}
Mona Baker and Gabriela Saldanha, editors. 2021.
\newblock \emph{Routledge encyclopedia of translation studies}, 3 edition.
\newblock Taylor \& Francis, London, England.

\bibitem[{Barzilay and McKeown(2001)}]{barzilay-mckeown-2001-extracting}
Regina Barzilay and Kathleen~R. McKeown. 2001.
\newblock \href {https://doi.org/10.3115/1073012.1073020} {Extracting
  paraphrases from a parallel corpus}.
\newblock In \emph{Proceedings of the 39th Annual Meeting of the Association
  for Computational Linguistics}, pages 50--57, Toulouse, France. Association
  for Computational Linguistics.

\bibitem[{Bawden et~al.(2018)Bawden, Sennrich, Birch, and
  Haddow}]{bawden-etal-2018-evaluating}
Rachel Bawden, Rico Sennrich, Alexandra Birch, and Barry Haddow. 2018.
\newblock \href {https://doi.org/10.18653/v1/N18-1118} {Evaluating discourse
  phenomena in neural machine translation}.
\newblock In \emph{Proceedings of the 2018 Conference of the North {A}merican
  Chapter of the Association for Computational Linguistics: Human Language
  Technologies, Volume 1 (Long Papers)}, pages 1304--1313, New Orleans,
  Louisiana. Association for Computational Linguistics.

\bibitem[{Brown et~al.(2020)Brown, Mann, Ryder, Subbiah, Kaplan, Dhariwal,
  Neelakantan, Shyam, Sastry, Askell et~al.}]{brown2020language}
Tom Brown, Benjamin Mann, Nick Ryder, Melanie Subbiah, Jared~D Kaplan, Prafulla
  Dhariwal, Arvind Neelakantan, Pranav Shyam, Girish Sastry, Amanda Askell,
  et~al. 2020.
\newblock Language models are few-shot learners.
\newblock \emph{Advances in neural information processing systems},
  33:1877--1901.

\bibitem[{Callison-Burch(2009)}]{callison-burch-2009-fast}
Chris Callison-Burch. 2009.
\newblock \href {https://aclanthology.org/D09-1030} {Fast, cheap, and creative:
  Evaluating translation quality using {A}mazon{'}s {M}echanical {T}urk}.
\newblock In \emph{Proceedings of the 2009 Conference on Empirical Methods in
  Natural Language Processing}, pages 286--295, Singapore. Association for
  Computational Linguistics.

\bibitem[{Castilho and Resende(2022)}]{castilho-2022}
Sheila Castilho and NatÃ¡lia Resende. 2022.
\newblock \href {https://doi.org/10.3390/info13020066} {Post-editese in
  literary translations}.
\newblock \emph{Information}, 13(2).

\bibitem[{Chakrabarty et~al.(2021)Chakrabarty, Saakyan, and
  Muresan}]{chakrabarty-etal-2021-dont}
Tuhin Chakrabarty, Arkadiy Saakyan, and Smaranda Muresan. 2021.
\newblock \href {https://doi.org/10.18653/v1/2021.emnlp-main.577} {Don{'}t go
  far off: An empirical study on neural poetry translation}.
\newblock In \emph{Proceedings of the 2021 Conference on Empirical Methods in
  Natural Language Processing}, pages 7253--7265, Online and Punta Cana,
  Dominican Republic. Association for Computational Linguistics.

\bibitem[{Chatterjee et~al.(2018)Chatterjee, Negri, Raphael, and
  Turchi}]{chatterjee2018findings}
Rajen Chatterjee, Matteo Negri, Rubino Raphael, and Marco Turchi. 2018.
\newblock Findings of the wmt 2018 shared task on automatic post-editing.
\newblock In \emph{Third Conference on Machine Translation (WMT)}, pages
  723--738. Association for Computational Linguistics (ACL).

\bibitem[{Chesterman(2005)}]{Chesterman2017-strat-papaers}
Andrew Chesterman. 2005.
\newblock Problems with strategies.
\newblock In K.~Károly and A.~Fóris, editors, \emph{Trends in Translation
  Studies. In honour of Kinga Klaudy.}, pages 17--28. Akademiai Kiado,
  Budapest.

\bibitem[{Fonteyne et~al.(2020)Fonteyne, Tezcan, and
  Macken}]{fonteyne_literary_2020}
Margot Fonteyne, Arda Tezcan, and Lieve Macken. 2020.
\newblock \href {https://aclanthology.org/2020.lrec-1.468} {Literary {Machine}
  {Translation} under the {Magnifying} {Glass}: {Assessing} the {Quality} of an
  {NMT}-{Translated} {Detective} {Novel} on {Document} {Level}}.
\newblock In \emph{Proceedings of the 12th {Language} {Resources} and
  {Evaluation} {Conference}}, pages 3790--3798, Marseille, France. European
  Language Resources Association.

\bibitem[{Genzel et~al.(2010)Genzel, Uszkoreit, and
  Och}]{genzel-etal-2010-poetic}
Dmitriy Genzel, Jakob Uszkoreit, and Franz Och. 2010.
\newblock \href {https://aclanthology.org/D10-1016} {{``}poetic{''} statistical
  machine translation: Rhyme and meter}.
\newblock In \emph{Proceedings of the 2010 Conference on Empirical Methods in
  Natural Language Processing}, pages 158--166, Cambridge, MA. Association for
  Computational Linguistics.

\bibitem[{Holtzman et~al.(2019)Holtzman, Buys, Du, Forbes, and
  Choi}]{holtzman2019curious}
Ari Holtzman, Jan Buys, Li~Du, Maxwell Forbes, and Yejin Choi. 2019.
\newblock The curious case of neural text degeneration.
\newblock \emph{arXiv preprint arXiv:1904.09751}.

\bibitem[{Jean et~al.(2017)Jean, Lauly, Firat, and Cho}]{Jean2017DoesNM}
S{\'e}bastien Jean, Stanislas Lauly, Orhan Firat, and Kyunghyun Cho. 2017.
\newblock Does neural machine translation benefit from larger context?
\newblock \emph{ArXiv}, abs/1704.05135.

\bibitem[{Jiang et~al.(2022)Jiang, Liu, Ma, Zhang, Yang, Huang, Sennrich,
  Cotterell, Sachan, and Zhou}]{jiang-etal-2022-blonde}
Yuchen Jiang, Tianyu Liu, Shuming Ma, Dongdong Zhang, Jian Yang, Haoyang Huang,
  Rico Sennrich, Ryan Cotterell, Mrinmaya Sachan, and Ming Zhou. 2022.
\newblock \href {https://doi.org/10.18653/v1/2022.naacl-main.111} {{BlonDe}: An
  automatic evaluation metric for document-level machine translation}.
\newblock In \emph{Proceedings of the 2022 Conference of the North American
  Chapter of the Association for Computational Linguistics: Human Language
  Technologies}, pages 1550--1565, Seattle, United States. Association for
  Computational Linguistics.

\bibitem[{Jones and Irvine(2013)}]{jones-irvine-2013-un}
Ruth Jones and Ann Irvine. 2013.
\newblock \href {https://aclanthology.org/W13-2713} {The (un)faithful machine
  translator}.
\newblock In \emph{Proceedings of the 7th Workshop on Language Technology for
  Cultural Heritage, Social Sciences, and Humanities}, pages 96--101, Sofia,
  Bulgaria. Association for Computational Linguistics.

\bibitem[{Junczys-Dowmunt(2019)}]{junczys-wmt}
Marcin Junczys-Dowmunt. 2019.
\newblock \href {https://doi.org/10.48550/ARXIV.1907.06170} {Microsoft
  translator at wmt 2019: Towards large-scale document-level neural machine
  translation}.

\bibitem[{Kasai et~al.(2021)Kasai, Sakaguchi, Bras, Dunagan, Morrison, Fabbri,
  Choi, and Smith}]{kasai2021bidimensional}
Jungo Kasai, Keisuke Sakaguchi, Ronan~Le Bras, Lavinia Dunagan, Jacob Morrison,
  Alexander~R Fabbri, Yejin Choi, and Noah~A Smith. 2021.
\newblock Bidimensional leaderboards: Generate and evaluate language hand in
  hand.
\newblock \emph{arXiv preprint arXiv:2112.04139}.

\bibitem[{Krippendorff(2011)}]{Krippendorff2011ComputingKA}
Klaus Krippendorff. 2011.
\newblock Computing krippendorff's alpha-reliability.

\bibitem[{Liu et~al.(2020)Liu, Gu, Goyal, Li, Edunov, Ghazvininejad, Lewis, and
  Zettlemoyer}]{liu2020multilingual}
Yinhan Liu, Jiatao Gu, Naman Goyal, Xian Li, Sergey Edunov, Marjan
  Ghazvininejad, Mike Lewis, and Luke Zettlemoyer. 2020.
\newblock Multilingual denoising pre-training for neural machine translation.
\newblock \emph{Transactions of the Association for Computational Linguistics},
  8:726--742.

\bibitem[{Macken et~al.({2022})Macken, Vanroy, Desmet, and
  Tezcan}]{macken-2022-pe}
Lieve Macken, Bram Vanroy, Luca Desmet, and Arda Tezcan. {2022}.
\newblock \href {{https://aclanthology.org/2022.eamt-1.13}} {{Literary
  translation as a three-stage process : machine translation, post-editing and
  revision}}.
\newblock In \emph{{Proceedings of the 23rd Annual Conference of the European
  Association for Machine Translation}}, pages {101--110}. {European
  Association for Machine Translation}.

\bibitem[{Miculicich et~al.(2018)Miculicich, Ram, Pappas, and
  Henderson}]{miculicich-etal-2018-document}
Lesly Miculicich, Dhananjay Ram, Nikolaos Pappas, and James Henderson. 2018.
\newblock \href {https://doi.org/10.18653/v1/D18-1325} {Document-level neural
  machine translation with hierarchical attention networks}.
\newblock In \emph{Proceedings of the 2018 Conference on Empirical Methods in
  Natural Language Processing}, pages 2947--2954, Brussels, Belgium.
  Association for Computational Linguistics.

\bibitem[{Molina and Hurtado~Albir(2004)}]{Molina2004-ms}
Luc{\'\i}a Molina and Amparo Hurtado~Albir. 2004.
\newblock Translation techniques revisited: A dynamic and functionalist
  approach.
\newblock \emph{Meta}, 47(4):498--512.

\bibitem[{Moorkens et~al.(2018)Moorkens, Toral, Castilho, and
  Way}]{moorkens2018translators}
Joss Moorkens, Antonio Toral, Sheila Castilho, and Andy Way. 2018.
\newblock Translators’ perceptions of literary post-editing using statistical
  and neural machine translation.
\newblock \emph{Translation Spaces}, 7(2):240--262.

\bibitem[{Needleman and Wunsch(1970)}]{needleman-wunsch-1970}
Saul~B. Needleman and Christian~D. Wunsch. 1970.
\newblock \href {https://doi.org/https://doi.org/10.1016/0022-2836(70)90057-4}
  {A general method applicable to the search for similarities in the amino acid
  sequence of two proteins}.
\newblock \emph{Journal of Molecular Biology}, 48(3):443--453.

\bibitem[{Neubert(1983)}]{neubert1983discourse}
Albrecht Neubert. 1983.
\newblock Discourse analysis of translation.

\bibitem[{Papineni et~al.(2002)Papineni, Roukos, Ward, and Zhu}]{bleu}
Kishore Papineni, Salim Roukos, Todd Ward, and Wei-Jing Zhu. 2002.
\newblock \href {https://doi.org/10.3115/1073083.1073135} {Bleu: A method for
  automatic evaluation of machine translation}.
\newblock In \emph{Proceedings of the 40th Annual Meeting on Association for
  Computational Linguistics}, ACL '02, page 311–318, USA. Association for
  Computational Linguistics.

\bibitem[{Sellam et~al.(2020)Sellam, Das, and Parikh}]{sellam-etal-2020-bleurt}
Thibault Sellam, Dipanjan Das, and Ankur Parikh. 2020.
\newblock \href {https://doi.org/10.18653/v1/2020.acl-main.704} {{BLEURT}:
  Learning robust metrics for text generation}.
\newblock In \emph{Proceedings of the 58th Annual Meeting of the Association
  for Computational Linguistics}, pages 7881--7892, Online. Association for
  Computational Linguistics.

\bibitem[{Taivalkoski-Shilov(2019)}]{taivalkoski2019free}
Kristiina Taivalkoski-Shilov. 2019.
\newblock Free indirect discourse: an insurmountable challenge for literary mt
  systems?
\newblock In \emph{Proceedings of the Qualities of Literary Machine
  Translation}, pages 35--39.

\bibitem[{Thompson and Post(2020)}]{thompson-post-2020-automatic}
Brian Thompson and Matt Post. 2020.
\newblock \href {https://doi.org/10.18653/v1/2020.emnlp-main.8} {Automatic
  machine translation evaluation in many languages via zero-shot paraphrasing}.
\newblock In \emph{Proceedings of the 2020 Conference on Empirical Methods in
  Natural Language Processing (EMNLP)}, pages 90--121, Online. Association for
  Computational Linguistics.

\bibitem[{Tiedemann and Scherrer(2017)}]{tiedemann-scherrer-2017-neural}
J{\"o}rg Tiedemann and Yves Scherrer. 2017.
\newblock \href {https://doi.org/10.18653/v1/W17-4811} {Neural machine
  translation with extended context}.
\newblock In \emph{Proceedings of the Third Workshop on Discourse in Machine
  Translation}, pages 82--92, Copenhagen, Denmark. Association for
  Computational Linguistics.

\bibitem[{Toral et~al.(2020)Toral, Oliver, and
  Ballest{\'\i}n}]{toral2020machine}
Antonio Toral, Antoni Oliver, and Pau~Ribas Ballest{\'\i}n. 2020.
\newblock Machine translation of novels in the age of transformer.
\newblock \emph{arXiv preprint arXiv:2011.14979}.

\bibitem[{Toral and Way(2015)}]{Toral2015TranslatingLT}
Antonio Toral and Andy Way. 2015.
\newblock Translating literary text between related languages using smt.
\newblock In \emph{CLfL@NAACL-HLT}.

\bibitem[{Toral and Way(2018)}]{toral2018level}
Antonio Toral and Andy Way. 2018.
\newblock What level of quality can neural machine translation attain on
  literary text?
\newblock In \emph{Translation Quality Assessment}, pages 263--287. Springer.

\bibitem[{Toral et~al.(2018)Toral, Wieling, and Way}]{toral2018post}
Antonio Toral, Martijn Wieling, and Andy Way. 2018.
\newblock Post-editing effort of a novel with statistical and neural machine
  translation.
\newblock \emph{Frontiers in Digital Humanities}, page~9.

\bibitem[{Voigt and Jurafsky(2012)}]{voigt2012towards}
Rob Voigt and Dan Jurafsky. 2012.
\newblock Towards a literary machine translation: The role of referential
  cohesion.
\newblock In \emph{Proceedings of the NAACL-HLT 2012 Workshop on Computational
  Linguistics for Literature}, pages 18--25.

\bibitem[{Voita et~al.(2019)Voita, Sennrich, and
  Titov}]{voita-etal-2019-context}
Elena Voita, Rico Sennrich, and Ivan Titov. 2019.
\newblock \href {https://doi.org/10.18653/v1/D19-1081} {Context-aware
  monolingual repair for neural machine translation}.
\newblock In \emph{Proceedings of the 2019 Conference on Empirical Methods in
  Natural Language Processing and the 9th International Joint Conference on
  Natural Language Processing (EMNLP-IJCNLP)}, pages 877--886, Hong Kong,
  China. Association for Computational Linguistics.

\bibitem[{Wieting et~al.(2019)Wieting, Berg-Kirkpatrick, Gimpel, and
  Neubig}]{wieting-etal-2019-beyond}
John Wieting, Taylor Berg-Kirkpatrick, Kevin Gimpel, and Graham Neubig. 2019.
\newblock \href {https://doi.org/10.18653/v1/P19-1427} {Beyond {BLEU}:training
  neural machine translation with semantic similarity}.
\newblock In \emph{Proceedings of the 57th Annual Meeting of the Association
  for Computational Linguistics}, pages 4344--4355, Florence, Italy.
  Association for Computational Linguistics.

\bibitem[{Wilcoxon(1945)}]{Wilcoxon1945}
Frank Wilcoxon. 1945.
\newblock \href {https://doi.org/10.2307/3001968} {Individual comparisons by
  ranking methods}.
\newblock \emph{Biometrics Bulletin}, 1(6):80.

\bibitem[{Zaidan and
  Callison-Burch(2011)}]{zaidan-callison-burch-2011-crowdsourcing}
Omar~F. Zaidan and Chris Callison-Burch. 2011.
\newblock \href {https://aclanthology.org/P11-1122} {Crowdsourcing translation:
  Professional quality from non-professionals}.
\newblock In \emph{Proceedings of the 49th Annual Meeting of the Association
  for Computational Linguistics: Human Language Technologies}, pages
  1220--1229, Portland, Oregon, USA. Association for Computational Linguistics.

\bibitem[{Zhai et~al.(2020)Zhai, Liu, Zhong, Illouz, and
  Vilnat}]{zhai-etal-2020-building}
Yuming Zhai, Lufei Liu, Xinyi Zhong, Gbariel Illouz, and Anne Vilnat. 2020.
\newblock \href {https://aclanthology.org/2020.lrec-1.496} {Building an
  {E}nglish-{C}hinese parallel corpus annotated with sub-sentential translation
  techniques}.
\newblock In \emph{Proceedings of the Twelfth Language Resources and Evaluation
  Conference}, pages 4024--4033, Marseille, France. European Language Resources
  Association.

\bibitem[{Zhai et~al.(2018)Zhai, Max, and Vilnat}]{zhai-etal-2018-construction}
Yuming Zhai, Aur{\'e}lien Max, and Anne Vilnat. 2018.
\newblock \href {https://aclanthology.org/W18-3814} {Construction of a
  multilingual corpus annotated with translation relations}.
\newblock In \emph{Proceedings of the First Workshop on Linguistic Resources
  for Natural Language Processing}, pages 102--111, Santa Fe, New Mexico, USA.
  Association for Computational Linguistics.

\bibitem[{Zhang et~al.(2018)Zhang, Luan, Sun, Zhai, Xu, Zhang, and
  Liu}]{zhang-etal-2018-improving}
Jiacheng Zhang, Huanbo Luan, Maosong Sun, Feifei Zhai, Jingfang Xu, Min Zhang,
  and Yang Liu. 2018.
\newblock \href {https://doi.org/10.18653/v1/D18-1049} {Improving the
  transformer translation model with document-level context}.
\newblock In \emph{Proceedings of the 2018 Conference on Empirical Methods in
  Natural Language Processing}, pages 533--542, Brussels, Belgium. Association
  for Computational Linguistics.

\bibitem[{Zhao et~al.(2019)Zhao, Peyrard, Liu, Gao, Meyer, and
  Eger}]{zhao2019moverscore}
Wei Zhao, Maxime Peyrard, Fei Liu, Yang Gao, Christian~M. Meyer, and Steffen
  Eger. 2019.
\newblock Moverscore: Text generation evaluating with contextualized embeddings
  and earth mover distance.
\newblock In \emph{Proceedings of the 2019 Conference on Empirical Methods in
  Natural Language Processing}, Hong Kong, China. Association for Computational
  Linguistics.

\end{thebibliography}
\bibliographystyle{bib/acl_natbib}

\appendix

\section*{Appendix}
\label{sec:appendix}

\begin{table*}[h]
    \centering
    \resizebox{\textwidth}{!}{\begin{tabular}{|l|l|l|l|l|l|l|l|}
    \hline
    Title & Author & Gender & Split & Source Lang & Pub Year & \# Trans \\ \hline
    A Confession & Leo Tolstoy & M & test                                               & \textit{ru} & 1882 & 2 \\ \hline
    Botchan & Natsume Soseki & M & test                                                 & \textit{ja} & 1906 & 2 \\ \hline
    Doctor Glass & Hjalmar Soderberg & M & test                                         & \textit{sv} & 1905 & 2 \\ \hline
    Dom Casmurro & Machado De Assis & M & test                                          & \textit{pt} & 1899 & 2 \\ \hline
    The Castle & Franz Kafka & M & test & \textit{de} & 1924 & 4 \\ \hline
    Chaka & Thomas Mofolo & M & test & \textit{st} & 1948 & 2 \\ \hline
    Envy & Yury Olesha & M & test                                                       & \textit{ru} & 1927 & 2 \\ \hline
    Fairytales Part 1 & Dahans Christian Andersen & M & test & \textit{da} & 1875 & 2-3 \\ \hline
    Gora & Rabindranath Tagore & M & test & \textit{bn} & 1941 & 2 \\ \hline
    Journey By Moonlight & Antal Szerb & M & test                                       & \textit{hu} & 1937 & 2 \\ \hline
    Kokoro & Natsume Soseki & M & test                                                  & \textit{ja} & 1914 & 2 \\ \hline
    Romance Of The Three Kingdoms 1 & Luo Guanzhong & M & test                          & \textit{zh} & 1399 & 2 \\ \hline
    Romance Of The Three Kingdoms 2 & Luo Guanzhong & M & test                          & \textit{zh} & 1399 & 2 \\ \hline
    The Adventures Of Captain Hatteras & Jules Verne & M & test                        & \textit{fr} & 1866 & 2 \\ \hline
    The Gentleman From San Francisco & Ivan  Bunin & M & test                          & \textit{ru} & 1915 & 3 \\ \hline
    The Little Prince & Antoine De Saint-Exupery & M & test                            & \textit{fr} & 1943 & 2 \\ \hline
    The Magic Mountain & Thomas Mann & M & test                                        & \textit{de} & 1924 & 2 \\ \hline
    The Trial & Franz Kafka & M & test                                                 & \textit{de} & 1925 & 4 \\ \hline
    War With The Newts & Karel Capek & M & test                                        & \textit{cs} & 1936 & 2 \\ \hline
    We & Yevgeny Zamyatin & M & test                                                   & \textit{ru} & 1920 & 5 \\ \hline
    The Sorrows of Young Werther & Johann Wolfgang Von Goethe & M & test               & \textit{de} & 1774 & 2 \\ \hline
    A Hero Of Our Time & Mikhail Lermontov & M & train                                 & \textit{ru} & 1840 & 2 \\ \hline
    A Raw Youth & Fyodor Dostoevsky & M & train                                        & \textit{ru} & 1875 & 2 \\ \hline
    Against The Grain & Joris Karl Huysmans & M & train                                & \textit{fr} & 1884 & 2 \\ \hline
    Amerika & Franz Kafka & M & train                                                  & \textit{de} & 1927 & 2 \\ \hline
    Anna Karenina & Leo Tolstoy & M & train                                            & \textit{ru} & 1878 & 2 \\ \hline
    Around The World In Eighty Days & Jules Verne & M & train                          & \textit{fr} & 1873 & 2 \\ \hline
    Beware Of Pity & Stefan Zweig & M & train                                          & \textit{de} & 1939 & 2 \\ \hline
    Brothers Karamazov & Fyodor Dostoevsky & M & train                                 & \textit{ru} & 1879 & 3 \\ \hline
    Buddenbrooks & Thomas Mann & M & train                                             & \textit{de} & 1901 & 2 \\ \hline
    Call To Arms & Lu Xun & M & train                                                  & \textit{zh} & 1923 & 2 \\ \hline
    Crime And Punishment & Fyodor Dostoevsky & M & train                               & \textit{ru} & 1866 & 3 \\ \hline
    Dead Souls & Nikolai Gogol & M & train                                             & \textit{ru} & 1842 & 4 \\ \hline
    Death In Venice & Thomas Mann & M & train                                          & \textit{de} & 1912 & 3 \\ \hline
    Demons & Fyodor Dostoevsky & M & train                                             & \textit{ru} & 1871 & 2 \\ \hline
    Don Quixote & Miguel De Cervantes & M & train                                      & \textit{es} & 1605 & 2 \\ \hline
    Elective Affinities & Johann Wolfgang Von Goethe & M & train                       & \textit{de} & 1809 & 2 \\ \hline
    Fairytales Part 2 & Dahans Christian Andersen & M & train & \textit{da} & 1875 & 2-3 \\ \hline
    Fathers And Sons & Ivan Turgenev & M & train                                       & \textit{ru} & 1862 & 3 \\ \hline
    Gargantua And Pantagruel & François Rabelais & M & train                           & \textit{fr} & 1532 & 2 \\ \hline
    Germinal & Emile Zola & M & train                                                  & \textit{fr} & 1885 & 2 \\ \hline
    Heidi & Johanna Spyri & F & train                                                  & \textit{de} & 1881 & 3 \\ \hline
    Hesitation & Lu Xun & M & train                                                    & \textit{zh} & 1926 & 2 \\ \hline
    Home Of The Gentry & Ivan Turgenev & M & train                                     & \textit{ru} & 1859 & 2 \\ \hline
    In A Grove & Ryunosuke Akutagawa & M & train                                       & \textit{ja} & 1922 & 2 \\ \hline
    In The Shadow Of Young Girls In Flower & Marcel Proust & M & train                 & \textit{fr} & 1918 & 2 \\ \hline
    Jacques The Fatalist & Denis Diderot & M & train                                   & \textit{fr} & 1796 & 2 \\ \hline
    Kallocain & Karin Boye & F & train                                                 & \textit{sv} & 1940 & 2 \\ \hline
    Kappa & Ryunosuke Akutagawa & M & train                                            & \textit{ja} & 1927 & 2 \\ \hline
    Kristin Lavransdatter 1 The Wreath & Sigrid Undset & F & train                     & \textit{nb} & 1920 & 2 \\ \hline
    Kristin Lavransdatter 2 The Wife & Sigrid Undset & F & train                       & \textit{nb} & 1920 & 2 \\ \hline
    Lassommoir & Emile Zola & M & train                                                & \textit{fr} & 1877 & 2 \\ \hline
    Les Miserables & Victor Hugo & M & train                                           & \textit{fr} & 1862 & 3 \\ \hline
    Manon Lescaut & Antoine François Prevost & M & train                               & \textit{fr} & 1731 & 2 \\ \hline
    Nana & Emile Zola & M & train                                                      & \textit{fr} & 1880 & 3 \\ \hline
    No Longer Human & Osamu Dazai & M & train                                          & \textit{ja} & 1948 & 2 \\ \hline
    Notes From Underground & Fyodor Dostoevsky & M & train                             & \textit{ru} & 1864 & 3 \\ \hline
    \end{tabular}}
\end{table*}

\begin{table*}[h]
    \centering
    \resizebox{\textwidth}{!}{
        \begin{tabular}{|l|l|l|l|l|l|l|l|}
        \hline
        Title & Author & Gender & Split & Source Lang & Pub Year & \# Trans \\ \hline
        Oblomov & Ivan Goncharov & M & train                                               & \textit{ru} & 1859 & 3 \\ \hline
        Petersburg & Andrei Bely & M & train                                               & \textit{ru} & 1913 & 3 \\ \hline
        Pinocchio & Carlo Collodi & M & train                                              & \textit{it} & 1883 & 2 \\ \hline
        Poor Folk & Fyodor Dostoevsky & M & train                                          & \textit{ru} & 1846 & 3 \\ \hline
        Rashomon & Ryunosuke Akutagawa & M & train                                         & \textit{ja} & 1915 & 3 \\ \hline
        Song Of The Little Road & Bibhutibhushan Bandyopadhyay & M & train & \textit{bn} & 1950 & 2 \\ \hline
        Steppenwolf & Hermann Hesse & M & train                                            & \textit{de} & 1927 & 2 \\ \hline
        Strange Tales From A Chinese Studio & Pu Songling & M & train                  & \textit{zh} & 1740 & 2 \\ \hline
        Swanns Way & Marcel Proust & M & train                                         & \textit{fr} & 1913 & 2 \\ \hline
        The Blind Owl & Sadegh Hedayat & M & train                                     & \textit{fa} & 1937 & 2 \\ \hline
        The Book Of Disquietude & Fernando Pessoa & M & train                          & \textit{pt} & 1982 & 2 \\ \hline
        The Count Of Monte Cristo & Alexandre Dumas & M & train                        & \textit{fr} & 1844 & 2 \\ \hline
        The Dancing Girl Of Izu & Yasunari Kawabata & M & train                        & \textit{ja} & 1926 & 2 \\ \hline
        The Death Of Ivan Ilyich & Leo Tolstoy & M & train                             & \textit{ru} & 1886 & 3 \\ \hline
        The Debacle & Emile Zola & M & train                                           & \textit{fr} & 1892 & 2 \\ \hline
        The Diary Of A Young Girl & Anne Frank & F & train                             & \textit{nl} & 1947 & 2 \\ \hline
        The Fortune Of The Rougons & Emile Zola & M & train                            & \textit{fr} & 1871 & 2 \\ \hline
        The Good Soldier Schweik 1 Behind The Lines & Jaroslav Hasek & M & train       & \textit{cs} & 1921 & 2 \\ \hline
        The Good Soldier Schweik 2 At The Front & Jaroslav Hasek & M & train           & \textit{cs} & 1922 & 2 \\ \hline
        The Good Soldier Schweik 3 The Glorious Licking & Jaroslav Hasek & M & train   & \textit{cs} & 1922 & 2 \\ \hline
        The Hunchback Of Notre Dame & Victor Hugo & M & train                          & \textit{fr} & 1833 & 2 \\ \hline
        The Journey To The West & Wu Cheng-En & M & train                              & \textit{zh} & 1592 & 4 \\ \hline
        The Kill & Emile Zola & M & train                                              & \textit{fr} & 1871 & 2 \\ \hline
        The Kreutzer Sonata & Leo Tolstoy & M & train                                  & \textit{ru} & 1889 & 2 \\ \hline
        The Manuscript Found In Saragossa & Jan Potocki & M & train                    & \textit{pl} & 1805 & 2 \\ \hline
        The Master And Margarita & Mikhail Bulgakov & M & train                        & \textit{ru} & 1966 & 2 \\ \hline
        The Mate Mattia Pascal & Luigi Pirandello & M & train                          & \textit{it} & 1904 & 2 \\ \hline
        The Metamorphosis & Franz Kafka & M & train                                    & \textit{de} & 1915 & 3 \\ \hline
        The Notebooks Of Malte Laurids Brigge & Rainer Maria Rilke & M & train         & \textit{de} & 1910 & 3 \\ \hline
        The Nun & Denis Diderot & M & train                                            & \textit{fr} & 1780 & 2 \\ \hline
        The Phantom Of The Opera & Gaston Leroux & M & train                           & \textit{fr} & 1909 & 3 \\ \hline
        The Posthumous Memoirs Of Bras Cubas & Joaquim Maria Machado De Assis & M & train & \textit{pt} & 1881 & 2 \\ \hline
        The Queen Of Spades & Alexander Pushkin & M & train                            & \textit{ru} & 1834 & 2 \\ \hline
        The Red And The Black & Stendhal & M & train                                   & \textit{fr} & 1830 & 2 \\ \hline
        The Story Of Gosta Berling & Selma Agerlof & F & train                         & \textit{sv} & 1891 & 2 \\ \hline
        The Three Musketeers & Alexandre Dumas & M & train                             & \textit{fr} & 1844 & 2 \\ \hline
        Twenty Thousand Leagues Under The Sea & Jules Verne & M & train                & \textit{fr} & 1869 & 3 \\ \hline
        Venus In Furs & Leopold Von Sacher-Masoch & M & train                          & \textit{de} & 1870 & 2 \\ \hline
        War And Peace & Leo Tolstoy & M & train                                        & \textit{ru} & 1865 & 2 \\ \hline
        Wild Geese & Mori Ogai & M & train                                             & \textit{ja} & 1911 & 2 \\ \hline
        Voyage Around My Room & Xavier De Maistre & M & valid                          & \textit{fr} & 1794 & 2 \\ \hline
        Bel Ami & Guy De Maupassant & M & valid                                        & \textit{fr} & 1885 & 2 \\ \hline
        Candide & Voltaire & M & valid                                                & \textit{fr} & 1759 & 2 \\ \hline
        Dream Of The Red Chamber & Cao Xueqin & M & valid                             & \textit{zh} & 1791 & 2 \\ \hline
        Dream Story & Arthur Schnitzler & M & valid                                   & \textit{de} & 1926 & 2 \\ \hline
        Kusamakura & Natsume Soseki & M & valid                                       & \textit{ja} & 1906 & 2 \\ \hline
        Madame Bovary & Gustave Flaubert & M & valid                                  & \textit{fr} & 1856 & 2 \\ \hline
        Ponniyin Selvan 1 The First Floods & Kalki Krishnamurthy & M & valid          & \textit{ta} & 1950 & 3 \\ \hline
        Siddhartha An Indian Tale & Hermann Hesse & M & valid                         & \textit{de} & 1922 & 2 \\ \hline
        The Alienist & Machado De Assis & M & valid                                   & \textit{pt} & 1881 & 2 \\ \hline
        The Captains Daughter & Aleksandr Pushkin & M & valid                         & \textit{ru} & 1836 & 2 \\ \hline
        The Idiot & Fyodor Dostoevsky & M & valid                                     & \textit{ru} & 1868 & 4 \\ \hline
        The Sin Of Abbe Mouret & Emile Zola & M & valid                               & \textit{fr} & 1875 & 2 \\ \hline
        The Twelve Chairs & Ilya Ilf And Yevgeny Petrov & M & valid & \textit{ru} & 1947 & 2 \\ \hline
        \end{tabular}
    }
\caption{A full list of the literary texts from which the source paragraphs in \name\ are sampled with author name, author gender, publication year, source language, and test/train/val split designations.}
\label{tab:par3_all_data}
\end{table*}

\begin{table*}[]
    \renewcommand{\arraystretch}{1.15}
    \centering
    \begin{tabular}{|l|c|c|c|}
        \hline
         & \texttt{HUM} $>$ \texttt{GTr}  & \texttt{GPT-3} $>$ \texttt{GTr} & \texttt{HUM} $>$ \texttt{GPT-3} \\ \hline
        French & 86.0\%{\small *} & 66.0\%{\small *} & 64.0\%{\small *} \\ \hline
        Russian & 80.0\%{\small *} & 68.0\%{\small *} & 64.0\%{\small *} \\ \hline
        German & 86.0\%{\small *} & 78.0\%{\small ***} & 62.0\%{\small *} \\ \hline
    \end{tabular}
    \caption{The preference of translators for all three evaluations: (1) \texttt{HUM} vs \texttt{GTr}, (2) \texttt{GPT-3} vs \texttt{GTr}, (3) \texttt{HUM} vs \texttt{GPT-3}, divided by the language of translator. The translators unanimously preferred (1) \texttt{HUM}, (2) \texttt{GPT-3}, and (3) \texttt{HUM}. The significance is reported for the following levels {\small *}$p$<.05, {\small **}$p$<.01, {\small ***}$p$<.001 (binomial test).}
    \label{tab:par3_human_eval}
\end{table*}

\begin{table*}[]
    \renewcommand{\arraystretch}{1.15}
    \centering
    \footnotesize
    \begin{tabular}{|l|p{11cm}|}
        \hline
        \textbf{Edit-Type} & \textbf{Example} \\ \hline

        \textbf{Corrected Mistranslation} &
        \texttt{GTr}: First put me in a coffin and bury me in the ground, \textbf{\color{purple}then give me my daughter}. \\
        &\texttt{GPT-3}: Put me in the grave first and bury me in the ground, \textbf{\color{teal}then give my daughter away}. \\
        &\\
        &\texttt{GTr}: I have only one thing left: to tell you everything, my unknown readers (now you are \textbf{\color{purple}so for me the same roads}, and close, and inaccessible - as he was at that moment). \\
        &\texttt{GPT-3}:  I have only one thing left: to tell you everything, my unknown readers (now you are \textbf{\color{teal}as dear to me}, as close and as unattainable as he was at that moment). \\
        \hline

        \textbf{Corrected Pronouns} & 
        \texttt{GTr}: I tasted in \textbf{\color{purple}his} arms the delights of paradise, which produced those torments of hell with which you see me devoured; (...). \\
        &\texttt{GPT-3}:  I tasted the delights of paradise in \textbf{\color{teal}her} arms, which produced the torments of hell that you see me devoured by. \\
        &\\
        &\texttt{GTr}: \textbf{\color{purple}It} calmed \textbf{\color{purple}her} down and her heart stopped beating. \\
        &\texttt{GPT-3}: This calmed \textbf{\color{teal}me} down, \textbf{\color{teal}my} heart stopped pounding. \\
        &\\ \hline

        \textbf{Corrected Named Entities} &
        \texttt{GTr}: (...) by Lebedev's nephew, known to readers, and secondly, by \textbf{\color{purple} Ippolit. Hippolyte} was a very young man, about seventeen, maybe eighteen, with an intelligent, but constantly irritated expression on his face, on which illness left terrible marks. \\ 
        &\texttt{GPT-3}: (...) by Lebedev's nephew, a young man known to the reader, and secondly, by \textbf{\color{teal}Ippolit. Ippolit} was a very young man, about seventeen, or even eighteen, with an intelligent but constantly irritated expression on his face, on which illness had left terrible traces. \\
        \\ \hline

        \textbf{Less Literal / Better Wording} &
        \texttt{GTr}: \textbf{\color{purple}To amuse himself, he employed himself at home as a laborer}, and he even tried to paint the attic with \textbf{\color{purple}a residue of color} that the painters had left behind. \\
        &\texttt{GPT-3}: \textbf{\color{teal}To distract himself, he did odd jobs around the house}, and even tried to paint the attic with \textbf{\color{teal}some paint} that the painters had left over. \\
        \\ \hline

    \end{tabular}
    \caption{Examples of the GPT-3-post-edit (\texttt{GPT-3}) model correcting translations by Google Translate (\texttt{GTr})}
    \label{tab:post_edit_examples}
\end{table*}

\begin{figure*}[h!]
    \centering
    \includegraphics[width=1\textwidth]{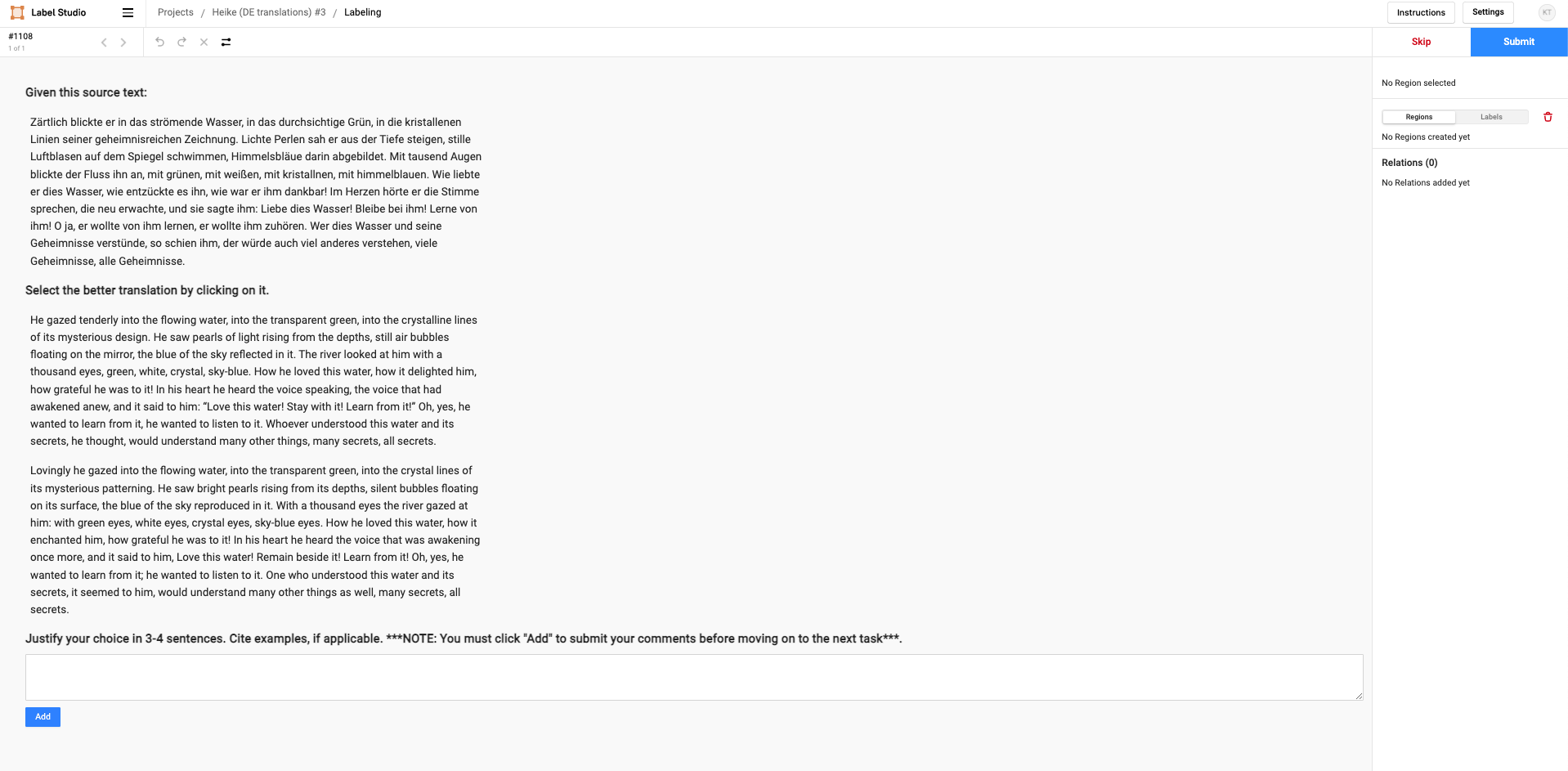}
    \caption{Example of the labeling interface.}
    \label{fig:labeling_example}
\end{figure*}

\begin{figure*}[h!]
    \centering
    \includegraphics[width=1\textwidth]{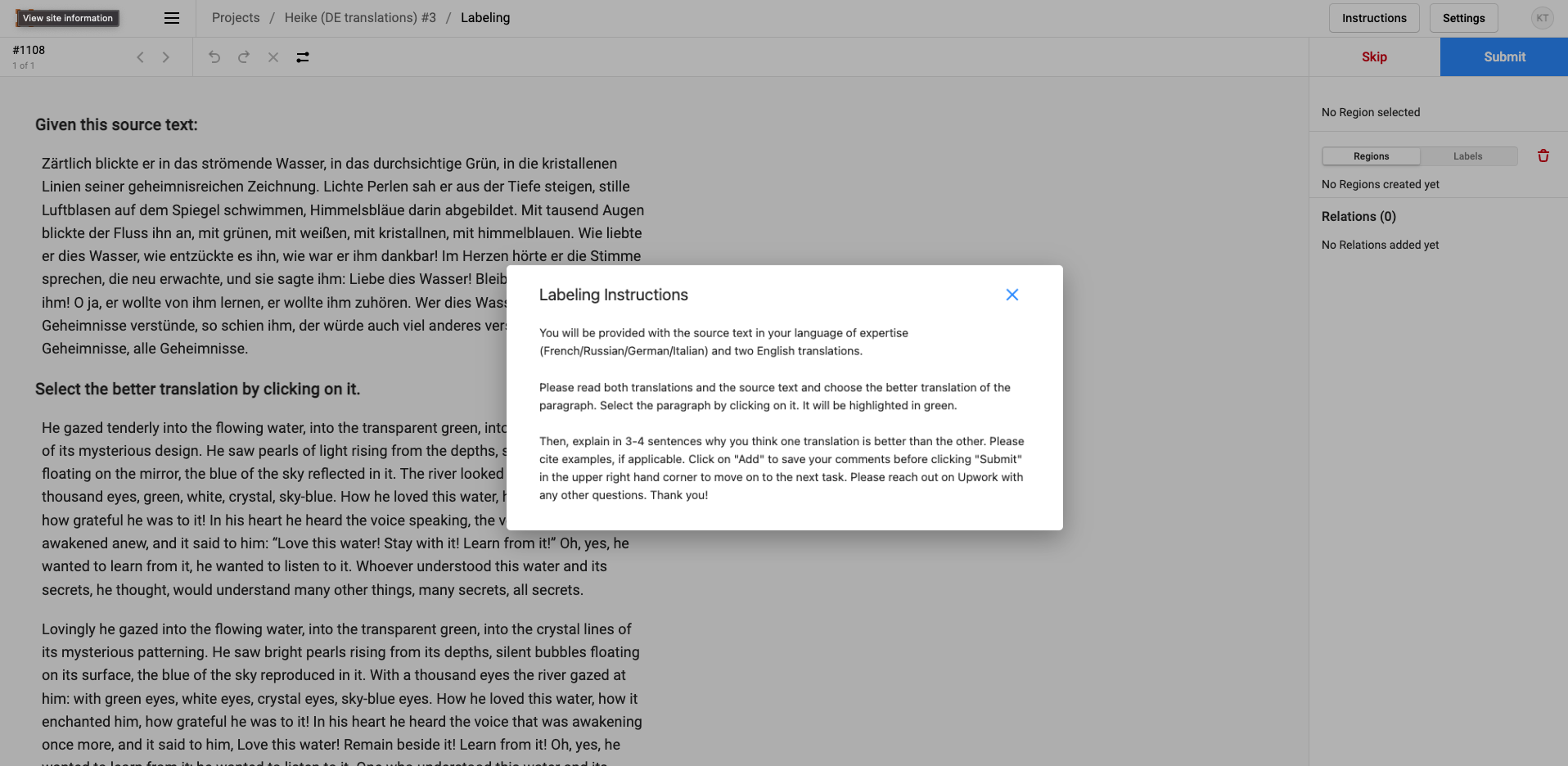}
    \caption{Example of the labeling instructions.}
    \label{fig:labeling_instructions}
\end{figure*}

\begin{figure*}[h!]
    \centering
    \includegraphics[width=0.5\textwidth]{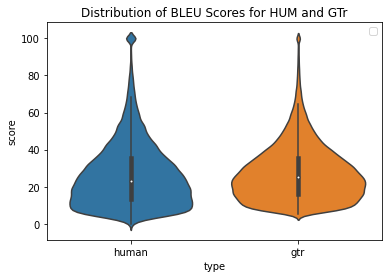}
    \caption{Distribution of \textsc{Bleu} scores for the \texttt{HUMAN} and \texttt{GTr} translations.}
    \label{fig:bleu_violin}
\end{figure*}

\begin{figure*}[h!]
    \centering
    \includegraphics[width=0.5\textwidth]{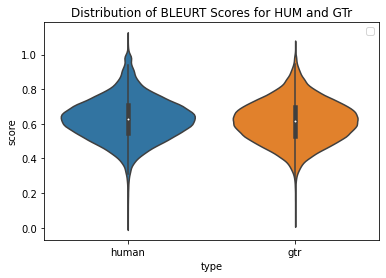}
    \caption{Distribution of \textsc{Bleurt} scores for the \texttt{HUMAN} and \texttt{GTr} translations.}
    \label{fig:bleurt_violin}
\end{figure*}

\begin{figure*}[h!]
    \centering
    \includegraphics[width=0.5\textwidth]{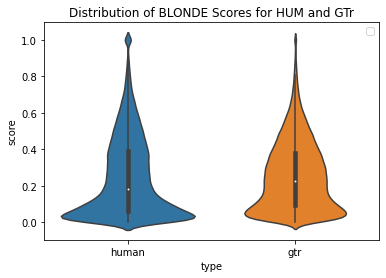}
    \caption{Distribution of \textsc{BlonDe} scores for the \texttt{HUMAN} and \texttt{GTr} translations.}
    \label{fig:blonde_violin}
\end{figure*}


\begin{table*}[!ht]
    \centering
    \caption{Examples of translators' techniques from \name. Catastrophic \texttt{GTr} mistakes were highlighted in {\color{red}red}.}
    \resizebox{\textwidth}{!}{\begin{tabular}{|p{3cm}|p{8cm}|p{16cm}|}
\hline
 
        \textbf{\large Technique} & \textbf{\large Description} & \textbf{\large Example} \\ 
 
        \hline
        \textbf{Borrowing} & Words or phrases left untranslated to introduce the \textit{flavor} of source language culture. & 
        \texttt{\textbf{SRC}}:\begin{otherlanguage}{russian}– Пиво есть? – сиплым голосом осведомился \textbf{\color{violet}Бездомный}.\end{otherlanguage}   (\emph{ru})
        
        \texttt{\textbf{HUM}}: “Got any beer?” inquired \textbf{\color{violet}Bezdomny} in a hoarse voice.

        \texttt{\textbf{GTr}}: - Do you have beer? \textbf{\color{violet}Homeless} inquired in a hoarse voice.

        (from \textit{Master and Margarita}) \\ 

\hline
        \textbf{Established Equivalence} & An equivalent of the source language using different stylistic and structural methods. This technique is applied frequently to idioms, clichés, simile, and proverbs. & \texttt{\textbf{SRC}}: \begin{CJK}{UTF8}{min}中学と師範とはどこの県下でも\textbf{\color{violet}犬と猿のように仲がわるい}そうだ。\end{CJK} (\emph{ja})

        \texttt{\textbf{HUM}}: The middle school and the normal, I understood, are as much \textbf{\color{violet}friendly as dogs and monkeys}. 

        \texttt{\textbf{GTr}}: It seems that junior high school and {\color{red}\textbf{instructors}} \textbf{\color{violet}get along with each other like dogs and monkeys} in any prefecture.

        (from \textit{Botchan}) \\ 

\hline
        \textbf{Transposition} & A change in grammatical category, such like word class, number, tense, etc.  & \texttt{SRC}: Et il reprit son carnet, biffant \textbf{\color{violet}avec le plus grand soin} les sommes qu'il venait de payer. (\emph{fr})

        \texttt{HUM}: And he took up his notebook, \textbf{\color{violet}carefully} crossing out the amounts he had just paid.

        \texttt{GTr}: And he went back to his notebook, crossing out \textbf{\color{violet}with the greatest care} the sums he had just paid.

        (from \textit{The Count of Monte Cristo}) \\ 

\hline
        \textbf{Modulation} & A shift in point of view, focus, cognitive category. & 
        \texttt{\textbf{SRC}}: Bei der Schnelligkeit ihres Wesens war ihr \textbf{\color{violet}nicht leicht} zu widersprechen. (\emph{de})

        \texttt{\textbf{HUM}}: Being so quick in her manner she was \textbf{\color{violet}hard} to contradict.

        \texttt{\textbf{GTr}}: Given the quickness of her nature, it was \textbf{\color{violet}not easy} to contradict her.
 
        (from \textit{Elective Affinities}) \\ 

\hline
        \textbf{Addition} & An addition of a new piece of information, which is not easily inferable from the source language. & 
        \texttt{\textbf{SRC}}: \begin{CJK}{UTF8}{min}清が物をくれる時には必ずおやじも兄も居ない時に限る。\end{CJK} (\emph{ja})

        \textbf{\texttt{HUM}}: When Kiyo gave me these presents she would always \textbf{\color{violet}be careful to choose} times when the old man and my brother were not around. 

        \texttt{\textbf{GTr}}: When {\color{red}\textbf{Qing}} gives me something, {\color{red}\textbf{I}} always do it only when my father and brother are not there.

        (from \textit{Botchan}) \\ 

\hline
        \textbf{Omission} & An omission of information present in the source language to the extent that it is not even easily inferable in the target language. & 
        \texttt{\textbf{SRC}}: \begin{CJK}{UTF8}{min}健全なる\textbf{\color{violet}男女}の河童よ\end{CJK} (\emph{ja})

        \texttt{\textbf{HUM}}: IF YOU ARE HEALTHY {\color{violet}\textbf{\_\_\_}} KAPPAS

        \texttt{\textbf{GTr}}: Healthy \textbf{\color{violet}male and female} kappa

        (from \textit{Kappa}) \\ 

\hline
        \textbf{Generalization} & A word or phrase translated into a more general one (hypernym). & \texttt{\textbf{SRC}}: \begin{CJK}{UTF8}{min}妹子是被\textbf{\color{violet}大哥}吃了，母亲知道没有，我可不得而知。\end{CJK} (\emph{zh})

        \texttt{\textbf{HUM}}: My sister was eaten by my \textbf{\color{violet}brother}, but I don't know whether Mother realized it or not.

        \texttt{\textbf{GTr}}: The sister was eaten by \textbf{\color{violet}the elder brother}, and whether the mother knew it or not, I don't know.

        (from \textit{Call to Arms}) \\ 

\hline
        \textbf{Particularization} & A word or phrase is translated into a more precise or concrete term (hyponym). & 
        \texttt{\textbf{SRC}}: (...) Andrea saisit la main du comte, la serra, \textbf{\color{violet}sauta} dans son phaéton et disparut. (\emph{fr})

        \texttt{\textbf{HUM}}: (...) Andrea seized his hand, pressed it, \textbf{\color{violet}leapt} into his phaeton and rode off.

        \texttt{\textbf{GTr}}: (...) Andrea seized the count's hand, squeezed it, \textbf{\color{violet}jumped} into his phaeton and disappeared.

        (from \textit{The Count of Monte Cristo}) \\ 
        
\hline
        \textbf{Adaptation} & Content is adapted to the target culture. It may include adapting the portrayed situation so that it is appropriate for the target culture (cultural substitution). & 
        \texttt{\textbf{SRC}}: \begin{CJK}{UTF8}{min}父はこの前の冬に帰って来た時ほど\textbf{\color{violet}将棋}を差したがらなくなった。\end{CJK} (\emph{ja})

        \texttt{\textbf{HUM}}: My father did not show as much interest in \textbf{\color{violet}chess} as he had done the previous winter.

        \texttt{\textbf{GTr}}: My dad was less reluctant to play \textbf{\color{violet}shogi} than when he came back last winter.

        (from \textit{Kokoro}) \\ 

\hline
        \textbf{Description} & A term or expression from the source language is described in text in the translation. & 
        \texttt{\textbf{SRC}}: \begin{CJK}{UTF8}{min}或時先生が例の通りさっさと海から上がって来て、いつもの場所に脱ぎ棄すてた\textbf{\color{violet}浴衣}を着ようとすると、どうした訳か、その浴衣に砂がいっぱい着いていた。\end{CJK} (\emph{ja})

        \texttt{\textbf{HUM}}: One day, however, after his usual swim, Sensei was about to put on his \textbf{\color{violet}summer dress} which he had left on the bench, when he noticed that the dress, for some reason, was covered with sand.

        \texttt{\textbf{GTr}}: At one point, as usual, the teacher came up from the sea and tried to put on the \textbf{\color{violet}yukata} that had been taken off and thrown away at the usual place, but for some reason, the yukata was full of sand.

        (from \textit{Kokoro}) \\ 
\hline
        \textbf{Sentence Diffusion} & The source sentence is being translated into two or more sentences in the translation. & 
        \texttt{\textbf{SRC}}: Prodal jsem tě, kamaráde, hanebně prodal. (\emph{cs})

        \texttt{\textbf{HUM}}: I’ve sold you, buddy. Shamefully sold you. 

        \texttt{\textbf{GTr}}: I sold you, my friend, I shamefully sold you.

        (from \textit{The Good Soldier Schweik}) \\ 
\hline
        \textbf{Sentence Merging} & Two or more sentences from the source langage are combined together into one sentence in the translation. & 
        \texttt{\textbf{SRC}}: \begin{CJK}{UTF8}{min}三年前の夏のことです。僕は人並みにリュック・サックを背負い、あの上高地の温泉宿から穂高山へ登ろうとしました。\end{CJK} (\emph{ja})

        \texttt{\textbf{HUM}}: One summer morning three years ago, I left an inn at Kamikōchi hot spring to climb Mt. Hodaka, with a rucksack on my back. 

        \texttt{\textbf{GTr}}: It was the summer three years ago. I carried a rucksack on my back like a person and tried to climb Mt. Hotaka from that hot spring inn in Kamikochi.

        (from \textit{Kappa}) \\ 
        
\hline
        \textbf{Reordering} & Information is moved from one place in the paragraph to another for better coherence in the target language. & 
        \texttt{SRC}: \begin{CJK}{UTF8}{min}私が先生と知り合いになったのは鎌倉である。\end{CJK} (\emph{ja})

        \texttt{HUM}: It was at Kamakura, \textbf{\color{violet}during the summer holidays}, that I first met Sensei.

        \texttt{GTr}: It was Kamakura that I got to know the teacher.

        (from \textit{Kokoro}) \\ 

\hline
    \end{tabular}}
    \label{tab:strategies}
\end{table*}



\section{Dataset Versions}
\label{sec: versions}
The first version of \name\ was created in April 2022 as described in Section 2. The post-edit model and all human evaluations were conducted on this version of the dataset, which can still be found at \url{https://github.com/katherinethai/par3/}. In October 2022, we expanded \name\ to include three additional languages: Bengali, Sesotho, and Danish, along with new books in Russian and German. Those texts were translated using the Google Translate API in September 2022. The remaining data processing steps were the same.

\section{Post-editing Details}

\paragraph{Automatic evaluation of post-edited texts:} We compute \textsc{Bleu}, \textsc{Bleurt}, and \textsc{BlonDe} on the outputs of the post-editing model and present the results by source language in Table \ref{tab:par3_test_metrics}. All 3 metrics show a clear preference for the human translations or the post-edited outputs of GPT-3.
\begin{table*}[]
    \renewcommand{\arraystretch}{1.15}
    \centering
    \footnotesize
    \begin{tabular}{c|ccc|ccc|ccc}
        \hline
        \multicolumn{1}{c|}{\multirow{2}{*}{Source lang}} & \multicolumn{3}{c|}{\textsc{Bleu}} & \multicolumn{3}{c|}{\textsc{Bleurt}} & \multicolumn{3}{c}{\textsc{Blonde}} \\ \cline{2-10} 
        \multicolumn{1}{c|}{} & \multicolumn{1}{c|}{Hum} & \multicolumn{1}{c|}{GPT-3} & \multicolumn{1}{c|}{GTr} & \multicolumn{1}{c|}{Hum} & \multicolumn{1}{c|}{GPT-3} & \multicolumn{1}{c|}{GTr} & \multicolumn{1}{c|}{Hum} & \multicolumn{1}{c|}{GPT-3} & \multicolumn{1}{c}{GTr} \\ \hline
        \textit{fr} & 20.0 & 27.2 & 26.1 & 0.641 & 0.681 & 0.658 & 24.7 & 27.7 & 29.3 \\
        \textit{ru} & 46.0 & 38.2 & 36.8 & 0.636 & 0.631 & 0.612 & 30.1 & 24.5 & 24.3 \\
        \textit{de} & 19.8 & 22.2 & 19.0 & 0.525 & 0.552 & 0.530 & 18.0 & 21.1 & 18.6 \\
        \textit{ja} & 11.4 & 9.5 & 6.9 & 0.545 & 0.514 & 0.457 & 12.7 & 11.1 & 8.5\\
        \textit{zh} & 2.4 & 4.6 & 3.6 & 0.324 & 0.351 & 0.310 & 3.2 & 4.3 & 3.7 \\
        \textit{cs} & 19.4 & 22.7 & 19.1 & 0.625 & 0.621 & 0.590 & 18.3 & 22.0 & 19.7 \\
        \textit{pt} & 28.9 & 32.4 & 25.3 & 0.643 & 0.636 & 0.590 & 28.9 & 30.8 & 25.8 \\
        \textit{sv} & 28.1 & 33.8 & 29.2 & 0.649 & 0.673 & 0.538 & 27.2 & 33.5 & 31.7 \\
        \textit{hu} & 22.3 & 25.1 & 16.9 & 0.613 & 0.628 & 0.581 & 22.2 & 22.3 & 16.0 \\
        \hline
        \textbf{All} & 21.2 & 23.3 & 20.6 & 0.564 & 0.580 & 0.549 & 20.0 & 21.0 & 19.6 \\
        Win \%* & 28.5\% & 49.5\% & 22.0\% & 30.9\% & 52.1\% & 17.0\% & 30.5\% & 40.5\% & 29.0\% \\
        \hline
    \end{tabular}%
    \caption{The percentage of cases in which the automatic MT metric ranks the human, GPT-3, or Google translations above the other two. *Note: There are 9,648 unique source paragraphs that were input to the post-editing model, but we exclude ties in the calculation of Win \%. The total number of ties was 340, 94, and 33, for \textsc{Bleu}, \textsc{Bleurt}, and \textsc{BlonDe} respectively.}
    \label{tab:par3_test_metrics}
\end{table*}

\begin{table*}[]
    \renewcommand{\arraystretch}{1.15}
    \centering
    \begin{tabular}{c|cc|cc|cc}
        \hline
        \multicolumn{1}{c|}{\multirow{2}{*}{Source Lang}} & \multicolumn{2}{c|}{\textsc{Prism}} & \multicolumn{2}{c|}{\textsc{Prism-QE}} & \multicolumn{2}{c}{\textsc{MoverScore}} \\ \cline{2-7} 
        \multicolumn{1}{c|}{} & \multicolumn{1}{c|}{\texttt{Hum}} & \multicolumn{1}{c|}{\texttt{GTr}} & \multicolumn{1}{c|}{\texttt{Hum}} & \multicolumn{1}{c|}{\texttt{GTr}} & \multicolumn{1}{c|}{\texttt{Hum}} & \multicolumn{1}{c}{\texttt{GTr}} \\ \hline
        \textit{fr} & -2.3329 & -2.1711 & -2.1812 & -1.0883 & 0.5976 & 0.5985 \\
        \textit{ru} & -2.2142 & -2.1532 & -2.1472 & -1.2995 & 0.6109 & 0.5997 \\
        \textit{de} & -2.5624 & -2.3874 & -2.4816 & -1.5152 & 0.5912 & 0.5922 \\
        \textit{ja} & -3.0987 & -3.2028 & -3.2498 & -2.0923 & 0.5468 & 0.5281 \\
        \textit{zh} & -4.3927 & -4.2472 & -4.3900 & -3.3711 & 0.5191 & 0.5211 \\
        \textit{cs} & -3.0720 & -2.5455 & -2.5142 & -1.3088 & 0.5515 & 0.5704 \\
        \textit{pt} & -2.8693 & -2.4973 & -2.4732 & -1.0264 & 0.5805 & 0.5827 \\
        \textit{no} & -2.3435 & -2.2936 & -2.3826 & -1.1298 & 0.5938 & 0.5897 \\
        \textit{sv} & -1.7067 & -1.5924 & -1.6552 & -1.0648 & 0.6443 & 0.6408 \\
        \textit{it} & -2.1974 & -2.1698 & -2.1216 & -1.0742 & 0.5869 & 0.5894 \\
        \textit{es} & -2.1496 & -2.2906 & -2.2182 & -1.1592 & 0.6170 & 0.5875 \\
        \textit{fa} & -2.9812 & -2.9559 & -4.3144 & -4.0303 & 0.5735 & 0.5596 \\
        \textit{hu} & -2.3005 & -2.3425 & -2.3417 & -1.3059 & 0.6008 & 0.5701 \\
        \textit{nl} & -2.3712 & -2.1936 & -2.3491 & -1.0664 & 0.6010 & 0.6074 \\
        \textit{pl} & -2.0920 & -2.5984 & -2.6809 & -1.3299 & 0.6219 & 0.5685 \\
        \textit{ta} & -3.6783 & -3.6200 & -4.5426 & -4.3912 & 0.5341 & 0.5361 \\
        \hline
        \textbf{All} & -2.4207 & -2.2985 & -2.3290 & -1.3275 & 0.5966 & 0.5928 \\
        Win \%* & 34.61\% & 65.39\% & 3.41\% & 96.59\% & 45.44\% & 54.56\% \\
        \hline
    \end{tabular}
    \caption{Results of \textsc{Prism}, \textsc{Prism-QE} and \textsc{MoverScore} on \name\ . Higher score is better for all metrics. Scores were calculated on the entirety of version one of the \name\ dataset across its 107,467 unique source paragraphs. Again, we exclude ties from the calculation of Human Win \%. The total number of ties was 80, 82, and 100 for \textsc{Prism}, \textsc{Prism-QE} \cite{thompson-post-2020-automatic}, and \textsc{MoverScore} \cite{zhao2019moverscore}, respectively.}
    \label{tab:par3_other_metrics}
\end{table*}

\begin{table*}[]
    \renewcommand{\arraystretch}{1.15}
    \centering
    \begin{tabular}{|l|c|}
        \hline
        \textbf{Metrics} & \textbf{Kendall Tau} \\ \hline
        BLEU   & 0.209{\small ***} \\ \hline 
        BLONDE & 0.120{\small ***} \\ \hline
        BLEURT & 0.262{\small ***}\\ \hline
    \end{tabular}
    \caption{Metrics correlation with human evaluation. Significant correlation at {\small ***}$p$<.001}
    \label{tab:metric_correlations}
\end{table*}

\begin{table*}[]
    \renewcommand{\arraystretch}{1.15}
    \centering
    \begin{tabular}{|l|l|c|c|}
        \hline
                & Type          &   Wilcoxon-Pratt Signed-Rank Test &  Effect Size{\small *}  \\
        \hline
        \textsc{Bleu}    & \texttt{HUM} vs \texttt{GTr}        & $z=4.093,  p<.001$ &  0.236 \\
                & \texttt{GPT-3} vs \texttt{GTr}             & $z=-7.256, p<.001$ & 0.419  \\
                & \texttt{HUM} vs \texttt{GPT-3}             & $z=-1.888, p=.059$ & 0.109  \\
        \hline
        \textsc{BlonDe}  & \texttt{HUM} vs \texttt{GTr}        & $z=1.423, p=.155$ & 0.082  \\
                & \texttt{GPT-3} vs \texttt{GTr}            & $z=-5.127, p<.001$ &  0.296  \\
                & \texttt{HUM} vs \texttt{GPT-3}      & $z=-3.027, p=.003$ &  0.175  \\
        \hline
        \textsc{Bleurt}  & \texttt{HUM} vs \texttt{GTr}         & $z=7.0612, p<.001$ & 0.408  \\
                & \texttt{GPT-3} vs \texttt{GTr}        & $z=-7.553, p<.001$ &  0.436 \\
                & \texttt{HUM} vs \texttt{GPT-3}      & $z=1.827, p=.068$ &  0.105 \\
        \hline
    \end{tabular}
    \caption{Results of the performance of automatic metrics on the 150 paragraphs used in human evaluation.\\ {\small (*The common interpretation of the effect size is the following: 0.10-<0.30 (small), 0.30-<0.50 (moderate), >=0.50 (large))}}
    \label{tab:metrics_on_humaneval}
\end{table*}

\begin{table*}[]
    \scriptsize
    \begin{tabular}{p{7.3cm}p{7.3cm}}
        \multicolumn{2}{p{15cm}}{\texttt{SRC:} Joachim ging, und es kam die »Mittagssuppe«: ein einfältig symbolischer Name für das, was kam! Denn Hans Castorp war nicht auf Krankenkost gesetzt, – warum auch hätte man ihn darauf setzen sollen? Krankenkost, schmale Kost war auf keine Art indiziert bei seinem Zustande. Er lag hier und zahlte den vollen Preis, und was man ihm bringt in der stehenden Ewigkeit dieser Stunde, das ist keine »Mittagssuppe«, es ist das sechsgängige Berghof-Diner ohne Abzug und in aller Ausführlichkeit, – am Alltage üppig, am Sonntage ein Gala-, Lust- und Parademahl, von einem europäisch erzogenen Chef in der Luxushotelküche der Anstalt bereitet. Die Saaltochter, deren Amt es war, die Bettlägrigen zu versorgen, brachte es ihm unter vernickelten Hohldeckeln und in leckeren Tiegeln; sie schob den Krankentisch, der sich eingefunden, dies einbeinige Wunder von Gleichgewichtskonstruktion, quer über sein Bett vor ihn hin, und Hans Castorp tafelte daran wie der Sohn des Schneiders am Tischlein deck dich.}
        \vspace{.05cm}
        \\
        \hline
        \vspace{.05cm}
        \texttt{GTr}: Joachim went, and "Lunchtime Soup" came: a simple symbolic name for what was coming! Because Hans Castorp was not put on sick food - why should he have been put on it? Sick diet, small fare, was in no way indicated in his condition. He lay here and paid the full price, and what is brought to him in the standing eternity of this hour is not a "lunchtime soup," it is the six-course Berghof dinner without deduction and in great detail - sumptuous in everyday life, closed on Sundays Gala, pleasure and parade meal, prepared by a European-educated chef in the luxury hotel kitchen of the institution. The maid, whose job it was to look after the bedridden, brought it to him under nickel-plated hollow lids and in delicious jars; She pushed the patient's table that appeared, this one-legged marvel of balanced construction, across his bed in front of him, and Hans Castorp ate at it like the tailor's son at the little table, cover yourself. & \vspace{.05cm}\texttt{HUM}: Joachim would leave, and the “midday soup” would arrive—soup was the simplified, symbolic name for what came. Because Hans Castorp was not on a restricted diet—why should he have been? A restricted diet, short commons, would hardly have been appropriate to his condition. There he lay, paying full price, and what they brought him at this hour of fixed eternity was “midday soup,” the six-course Berghof dinner in all its splendor, with nothing missing—a hearty meal six days a week, a sumptuous showpiece, a gala banquet, prepared by a trained European chef in the sanatorium’s deluxe hotel kitchen. The dining attendant whose job it was to care for bedridden patients would bring it to him, a series of tasty dishes arranged under domed nickel covers. She would shove over the bed table, which was now part of the furniture, a marvel of one-legged equilibrium, adjust it across his bed in front of him, and Hans Castorp would dine from it like the tailor’s son who dined from a magic table.
        \vspace{.05cm}

    \end{tabular}

    \caption{An example \texttt{SRC} from Thomas Mann's \textit{The Magic Mountain} that was administered as an A/B test with its corresponding \texttt{GTr} and \texttt{HUM}. Though all monolingual raters chose \texttt{HUM}, the translator chose \texttt{GTr}.}
    \label{tab:hum_v_gt_1}
\end{table*}

\begin{table*}[]
    \scriptsize
    \begin{tabular}{p{7.3cm}p{7.3cm}}
        \multicolumn{2}{p{15cm}}{\texttt{SRC:} \begin{otherlanguage}{russian}Еще вначале, как только князь вошел в гостиную, он сел как можно дальше от китайской вазы, которою так напугала его Аглая. Можно ли поверить, что после вчерашних слов Аглаи в него вселилось какое-то неизгладимое убеждение, какое-то удивительное и невозможное предчувствие, что он непременно и завтра же разобьет эту вазу, как бы ни сторонился от нее, как бы ни избегал беды? Но это было так. В продолжение вечера другие сильные, но светлые впечатления стали наплывать в его душу; мы уже говорили об этом. Он забыл свое предчувствие. Когда он услышал о Павлищеве и Иван Федорович подвел и показал его снова Ивану Петровичу, он пересел ближе к столу и прямо попал на кресло подле огромной, прекрасной китайской вазы, стоявшей на пьедестале, почти рядом с его локтем, чуть-чуть позади. \end{otherlanguage}}
        \vspace{.05cm}
        \\
        \hline
        \vspace{.05cm}
        \texttt{GTr}: Even at the beginning, as soon as the prince entered the drawing room, he sat down as far as possible from the Chinese vase, with which Aglaya had so frightened him. Is it possible to believe that after Aglaya's words yesterday, some indelible conviction came into him, some amazing and impossible premonition that he would certainly break this vase tomorrow, no matter how he avoided it, no matter how he avoided trouble? But it was. In the course of the evening other strong but bright impressions began to flood into his soul; we already talked about this. He forgot his premonition. When he heard about Pavlishchev and Ivan Fyodorovich let him down and showed him again to Ivan Petrovich, he moved closer to the table and fell straight into an armchair beside a huge, beautiful Chinese vase, which stood on a pedestal, almost next to his elbow, a little behind. & \vspace{.05cm}\texttt{HUM}: From the very beginning, as soon as the prince entered the drawing room, he sat down as far as possible from the Chinese vase, with which Aglaya had frightened him so. Can one possibly believe that, after Aglaya’s words the day before, some sort of indelible conviction settled in him, some sort of astonishing and impossible premonition that the next day he would unfailingly break that vase, however far away he kept from it, however much he avoided the disaster? But it was so. In the course of the evening other strong but bright impressions began to flow into his soul; we have already spoken of that. He forgot his premonition. When he heard about Pavlishchev, and Ivan Fyodorovich brought him and introduced him again to Ivan Petrovich, he moved closer to the table and ended up right in the armchair next to the enormous, beautiful Chinese vase, which stood on a pedestal almost at his elbow, slightly behind him.

    \end{tabular}

    \caption{An example \texttt{SRC} from Fyodor Dostoevsky's \textit{The Idiot} that was administered as an A/B test with its corresponding \texttt{GTr} and \texttt{HUM}. Though all monolingual raters chose \texttt{HUM}, the translator chose \texttt{GTr}.}
    \label{tab:hum_v_gt_2}
\end{table*}

\begin{table*}[]
    \scriptsize
    \begin{tabular}{p{7.3cm}p{7.3cm}}
        \multicolumn{2}{p{15cm}}{\texttt{SRC:} \begin{otherlanguage}{russian}Князь, однако же, слышал, как его назвали идиотом, и вздрогнул, но не оттого, что его назвали идиотом. «Идиота» он тотчас забыл. Но в толпе, недалеко от того места, где он сидел, откуда-то сбоку — он бы никак не указал, в каком именно месте и в какой точке, — мелькнуло одно лицо, бледное лицо, с курчавыми темными волосами, с знакомыми, очень знакомыми улыбкой и взглядом, — мелькнуло и исчезло. Очень могло быть, что это только вообразилось ему; от всего видения остались у него в впечатлении кривая улыбка, глаза и светло-зеленый франтовской шейный галстук, бывший на промелькнувшем господине. Исчез ли этот господин в толпе или прошмыгнул в вокзал, князь тоже не мог бы определить.\end{otherlanguage}}
        \vspace{.05cm}
        \\
        \hline
        \vspace{.05cm}
        \texttt{GTr}: The prince, however, heard how he was called an idiot, and shuddered, but not because he was called an idiot. "Idiot" he immediately forgot. But in the crowd, not far from the place where he was sitting, from somewhere on the side - he would not have indicated exactly in what place and at what point - one face flashed, a pale face, with curly dark hair, with acquaintances, very familiar smile and look, flashed and disappeared. It could very well be that it was only his imagination; from the whole vision, he was impressed by the crooked smile, eyes and light green dandy neck tie that the gentleman flashed by. Whether this gentleman disappeared into the crowd or slipped into the station, the prince could not determine either. & \vspace{.05cm}\texttt{GPT-3}: The prince, however, heard how he was called an idiot, and he shuddered, but not because he was called an idiot. “Idiot” he immediately forgot. But in the crowd, not far from the place where he was sitting, from somewhere on the side—he would not have been able to indicate exactly where and in what place—a face flashed, a pale face, with curly dark hair, with a familiar, very familiar smile and gaze, flashed and disappeared. It could very well have been that it was only his imagination; from the whole vision he retained an impression of a crooked smile, eyes, and a light green necktie of the dandy who had flashed by. Whether this dandy disappeared into the crowd or slipped into the station, the prince would also not have been able to say.

    \end{tabular}

    \caption{An example \texttt{SRC} from Fyodor Dostoevsky's \textit{The Idiot} that was administered as an A/B test with its corresponding \texttt{GTr} and \texttt{GPT-3}. The translator preferred \texttt{GPT-3}.}
    \label{tab:gpt3_v_gt}
\end{table*}

\section{GPT-3 fine-tuning configuration for post-editing:}
\label{sec:config}
The model was fine-tuned on OpenAI's servers for 2 epochs, with a batch size of 32, a learning rate multiplier of 0.2, and a weight of 0.1 for loss on the prompt tokens. The finetuning took 3 hours total and cost \$565. Decoding on 9,648 test set examples\footnote{Some test set examples exceeded \textit{davinci}'s input limits.} was performed using nucleus sampling~\citep{holtzman2019curious} with $p=0.2$.\footnote{We performed a small-scale qualitative validation experiment on different values of $p$ to determine this hyperparameter.}

\end{document}